\documentclass[lettersize,journal]{IEEEtran}
\usepackage{amsmath,amsfonts}
\usepackage{algorithmic}
\usepackage{algorithm}
\usepackage{array}
\usepackage[caption=false,font=normalsize,labelfont=sf,textfont=sf]{subfig}
\usepackage{textcomp}
\usepackage{stfloats}
\usepackage{url}
\usepackage{verbatim}
\usepackage{graphicx}
\usepackage{cite}
\usepackage{amssymb}
\usepackage{booktabs}
\usepackage{multirow}

\graphicspath{{./figures/}}

\begin{document}

\title{ESAFusion: LiDAR--4-D Radar Fusion via Local Geometric Complementation and Multiscale Adaptive Interaction\\for 3-D Object Detection}

% \author{IEEE Publication Technology,~\IEEEmembership{Staff,~IEEE,}
% \author{IEEE Publication Technology,~\IEEEmembership{Staff,~IEEE,}
\author{Gang Ma, Senjie Hu, Junjie Liu, Chao Wang and Hui Wei
        % <-this % stops a space
% \thanks{This paper was produced by the IEEE Publication Technology Group. They are in Piscataway, NJ.}% <-this % stops a space
\thanks{Manuscript received September 24, 2026. \textit{(Corresponding author: Gang Ma.)}}

\thanks{ Gang Ma, Senjie Hu, Junjie Liu and Chao Wang are with the School of Future Technology, Shanghai University, Shanghai 200444, China (email: magang@shu.edu.cn; senjie\_hu@shu.edu.cn; liujunjie@shu.edu.cn; cwang@shu.edu.cn). 
		% (Authors' names and affiliation) First A. Author1 and Second B. Author2 are with the xxx Department, University of xxx, City, Zip code, Country, on leave from the National Institute for xxx, City, Zip code, Country (e-mail: author@domain.com). 

Hui Wei is with the Laboratory of Algorithms for Cognitive Models, School of Computer Science, Fudan University, Shanghai 200437, China (e-mail: weihui@fudan.edu.cn).	
            
            % Yan Peng is with the School of Future Technology, Shanghai University, Shanghai 200444, China (email:pengyan@shu.edu.cn). 
            }
}

% The paper headers
\markboth{Journal of \LaTeX\ Class Files,~Vol.~14, No.~8, August~2021}%
{Shell \MakeLowercase{\textit{et al.}}: A Sample Article Using IEEEtran.cls for IEEE Journals}

\IEEEpubid{0000--0000/00\$00.00~\copyright~2021 IEEE}
% Remember, if you use this you must call \IEEEpubidadjcol in the second
% column for its text to clear the IEEEpubid mark.

\maketitle

\begin{abstract}
LiDAR--4-D radar fusion combines accurate spatial geometry with motion and reflectivity cues from radar, offering a promising solution for 3-D object detection in complex driving environments.
However, sparse radar observations and differences in spatial sampling between the two modalities complicate reliable cross-modal complementation.
Moreover, the relative importance of modalities and feature scales varies across spatial regions, making adaptive fusion challenging.
To address these challenges, we propose ESAFusion, an evidence-aware and scale-adaptive framework that combines local geometric complementation with multiscale adaptive interaction.
Specifically, we introduce an Evidence-Aware Radar Selection (ERS) module to suppress radar clutter using motion and observation-quality evidence while retaining foreground confidence for subsequent fusion.
Then, the Pillar-Level Complementary Encoder (PCE) improves cross-modal complementation under mismatched spatial sampling using local geometric support from neighboring LiDAR pillars.
We further design an Intra- and Inter-Scale Adaptive Fusion (ISAF) module to adaptively adjust the contributions of different modalities and feature scales in bird's-eye-view (BEV) space.
Extensive experiments on the View-of-Delft (VoD) dataset show that ESAFusion achieves the highest mean average precision (mAP) among the compared methods, reaching 74.60\% in the Entire Annotated Area and 88.89\% in the Driving Corridor.
It also attains the highest average precision (AP) for Cyclist among these methods in both regions while running at 19.23~FPS.
Evaluation on the Astyx HiRes2019 dataset further validates the effectiveness of ESAFusion on an additional dataset, while experiments on VoD-Fog demonstrate robustness under progressively degraded LiDAR observations. The source code will be made publicly available at \url{https://github.com/SenJieHu549/ESAFusion}.
\end{abstract}


\section{Introduction}
% \IEEEPARstart{T}{his} 

\begin{figure}[!t]
    \centering
    \includegraphics[
        width=\columnwidth,
        trim=12bp 2bp 12bp 6bp,
        clip
    ]{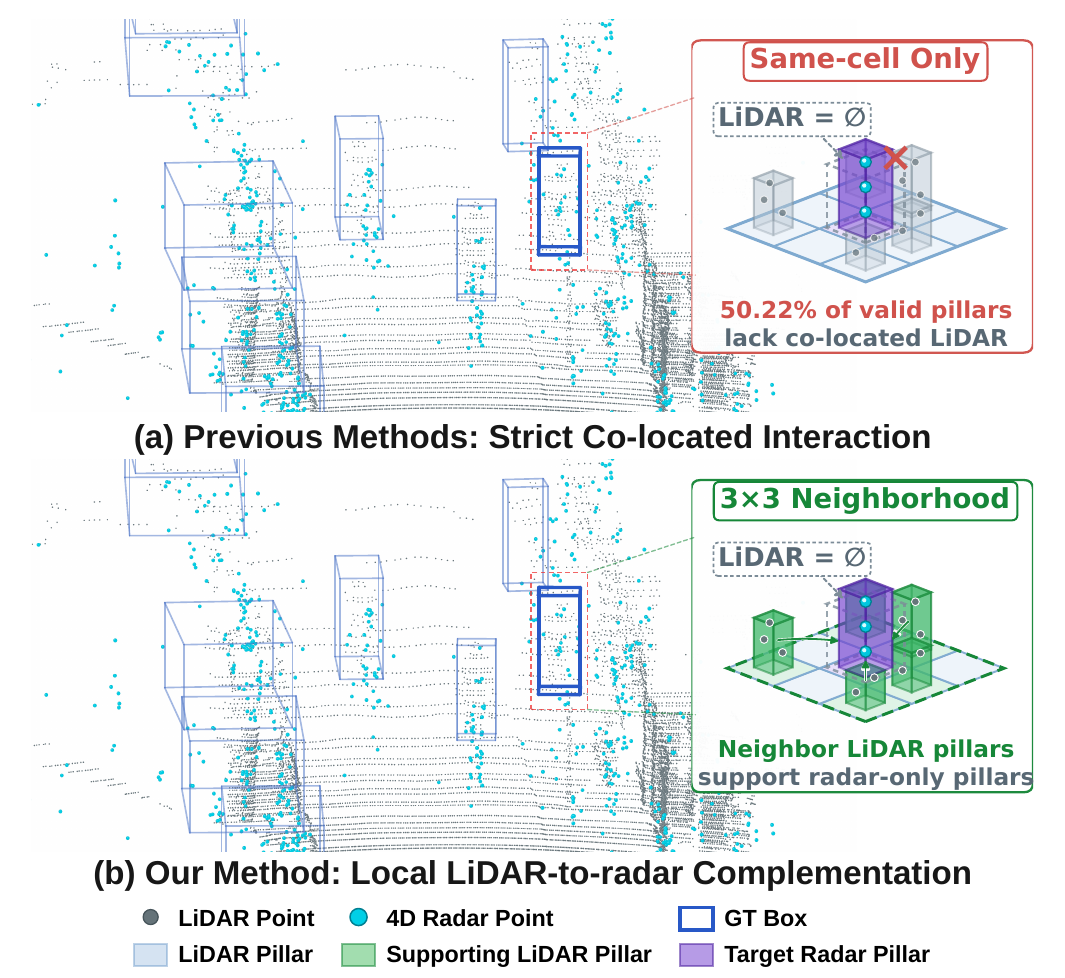}
   \caption{Comparison between strictly co-located fusion and ESAFusion. (a) Strict co-location leaves radar pillars without LiDAR support when same-cell LiDAR observations are absent. (b) ESAFusion complements such radar pillars using neighboring LiDAR observations within a local \(3\times3\) BEV neighborhood.}
    \label{fig:motivation}
\end{figure}

\IEEEPARstart{T}{hree-dimensional} object detection is fundamental to autonomous driving, particularly in complex scenes that demand robust perception~\cite{tong2026rel}.
LiDAR is widely used for its accurate geometric measurements and detailed structural information~\cite{qian2022survey,yang2026svefusion}.
However, owing to its inherent sampling characteristics, fewer points are captured from objects farther from the sensor, thereby degrading far-range detection performance~\cite{li2022sienet}.
Point-cloud incompleteness and occlusion further complicate reliable detection~\cite{lu2024hrnet}, while adverse weather degrades LiDAR measurements~\cite{huang2025l4dr,chae2025dopplerfusion}.
In contrast, 4-D radar provides robust long-range sensing under adverse weather conditions, together with Doppler velocity and radar cross section (RCS), which offer motion and reflectivity cues~\cite{fan2024radarsurvey,palffy2022vod,peng2025radaradverse}.
Nevertheless, 4-D radar point clouds are generally sparse, noisy, and irregularly distributed, limiting the detection performance of radar-only methods~\cite{liu2024smurf,bi2025maffnet}.
These complementary sensing characteristics have motivated increasing interest in LiDAR--4-D radar fusion for robust 3-D object detection~\cite{huang2025l4dr,yang2026svefusion,tong2026rel}.

\IEEEpubidadjcol

However, effective LiDAR--4-D radar fusion remains challenging across the point, pillar, and bird's-eye-view (BEV) levels.
At the point and pillar levels, differences in sampling density, measurement uncertainty, and physical attributes between LiDAR and radar complicate front-end cross-modal complementation~\cite{wang2023m2fusion,deng2024cmfa}.
Noisy radar observations introduce unreliable feature responses~\cite{bi2025maffnet}, while effectively exploiting radar-specific motion and reflectivity cues requires tailored feature encoding~\cite{chae2025dopplerfusion,huang2025l4dr}.
Meanwhile, mismatched spatial sampling leads to LiDAR and radar measurements of the same object falling into different BEV pillars.
As illustrated in Fig.~\ref{fig:motivation}(a), strictly co-located pillar interaction leaves radar pillars without LiDAR geometric support when co-located LiDAR observations are absent~\cite{huang2025l4dr}.
Nevertheless, as shown in Fig.~\ref{fig:motivation}(b), neighboring LiDAR pillars provide local geometric support for these radar pillars, motivating LiDAR-to-radar complementation beyond strict co-location.

\begin{figure}[!t]
    \centering
    \includegraphics[
        width=\columnwidth,
        trim={15bp 30bp 45bp 4bp},
        clip
    ]{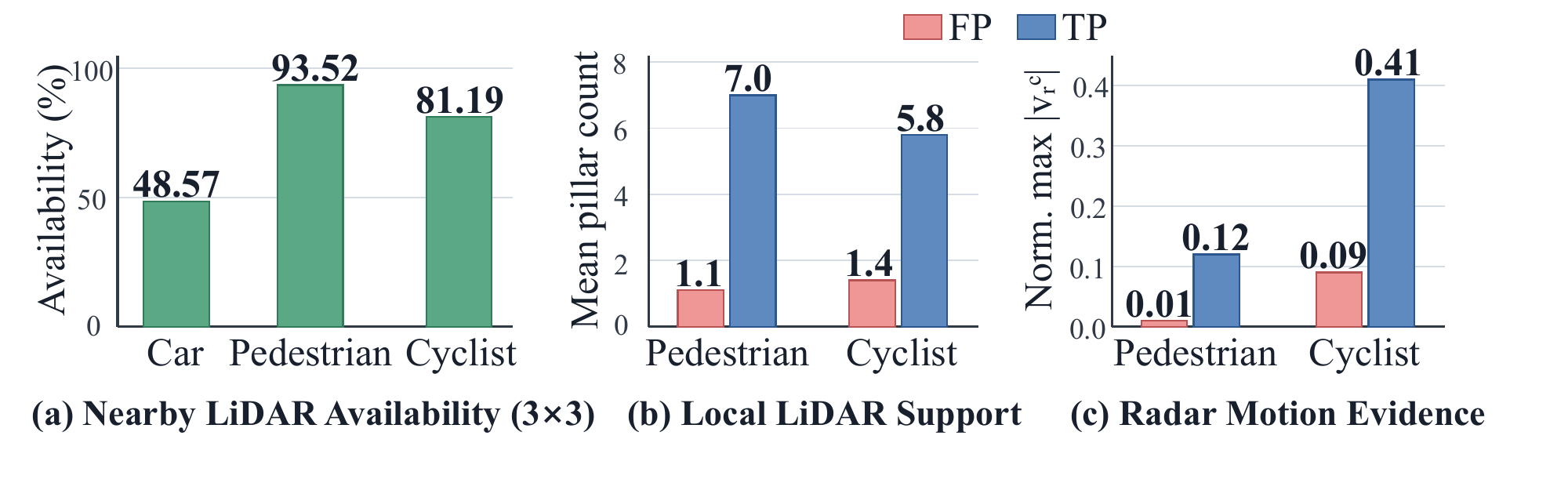}
    \caption{Quantitative analysis on the VoD validation set. (a) Class-wise availability of neighboring LiDAR observations. (b) Local LiDAR support for false-positive and true-positive detections. (c) Radar motion evidence for false-positive and true-positive detections.}
    \label{fig:quantitative_analysis}
\end{figure}

Quantitative analysis on the VoD validation set~\cite{palffy2022vod} further confirms that a frame-averaged 50.22\% of valid radar pillars lack co-located LiDAR support.
Among unsupported radar pillars within the corresponding class-specific ground-truth regions, 48.57\%, 93.52\%, and 81.19\% of Car, Pedestrian, and Cyclist pillars, respectively, have LiDAR observations within a local \(3\times3\) neighborhood, as shown in Fig.~\ref{fig:quantitative_analysis}(a).
Fig.~\ref{fig:quantitative_analysis}(b) and Fig.~\ref{fig:quantitative_analysis}(c) further show that radar pillars associated with true-positive detections exhibit higher local-support counts and stronger compensated radial-velocity responses than those associated with false positives.
Together, these observations provide quantitative support for evidence-aware radar selection and local geometric complementation.

At the BEV level, the relative importance of each modality and feature scale varies across spatial regions, reflecting differences in semantic content and local observation support.
High-resolution features preserve fine spatial details, whereas low-resolution features encode broader spatial context and higher-level semantic information~\cite{lin2017fpn}.
Representative fusion approaches use gating or attention mechanisms for modality interaction and incorporate features from multiple scales~\cite{wang2023m2fusion,huang2025l4dr}.
Despite these advances, joint adaptation of modality and scale responses to region-specific semantic content and local observation support remains underexplored.

To address these challenges, we propose ESAFusion, an evidence-aware and scale-adaptive framework for LiDAR--4-D radar fusion.
The Evidence-Aware Radar Selection (ERS) module first refines radar observations using motion and observation-quality evidence, filtering unreliable radar points while retaining foreground confidence for subsequent fusion.
The Pillar-Level Complementary Encoder (PCE) then extends LiDAR-to-radar geometric complementation from strictly co-located pillars to local neighborhoods, allowing radar pillars without co-located LiDAR observations to exploit nearby geometric support.
At the BEV level, the Intra- and Inter-Scale Adaptive Fusion (ISAF) module combines intra-scale modality gating with inter-scale feature recalibration.
Propagated radar evidence further refines radar gating, while multiscale semantic content and local observation support guide recalibration across scales.

The main contributions are summarized as follows:

\begin{enumerate}
    \item We develop an evidence-aware front end comprising ERS and PCE to address unreliable radar observations and mismatched spatial sampling. ERS filters unreliable radar points while retaining foreground confidence, whereas PCE extends LiDAR-to-radar geometric complementation beyond strict co-location using local LiDAR support based on geometric consistency and radar foreground evidence.
    \item We propose ISAF for joint modality and scale adaptation in BEV space. It combines intra-scale modality gating with inter-scale feature recalibration and incorporates radar evidence, multiscale semantic content, and local observation support to guide spatially varying fusion.
    \item Extensive experiments demonstrate that ESAFusion achieves strong detection performance on VoD and Astyx datasets, supports real-time inference, and maintains robustness on VoD-Fog under progressively degraded LiDAR observations.
\end{enumerate}

\begin{figure*}[!t]
    \centering
    \includegraphics[
        width=\textwidth,
        trim=16bp 117bp 4bp 70bp,
        clip
    ]{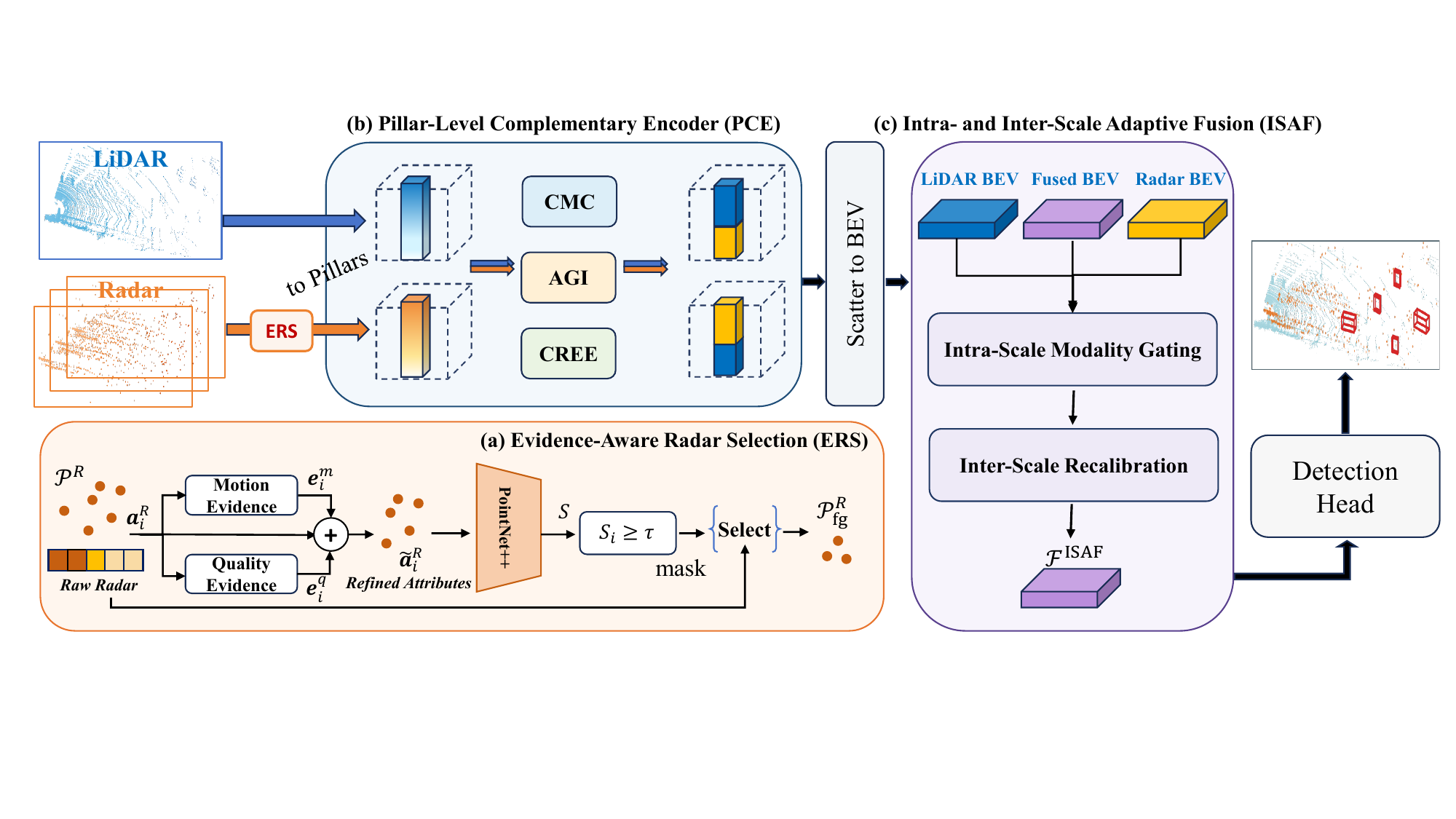}
    \caption{Overview of the ESAFusion framework. (a) Evidence-Aware Radar Selection (ERS) models motion and observation-quality evidence to refine radar attributes and suppress low-confidence clutter. (b) Pillar-Level Complementary Encoder (PCE) performs co-located attribute complementation, incorporates geometrically consistent local LiDAR support, and enhances radar pillars with compact radar evidence. (c) Intra- and Inter-Scale Adaptive Fusion (ISAF) adaptively gates LiDAR and radar features within each scale and performs support-conditioned inter-scale recalibration to produce a fused BEV representation for detection.}
    \label{fig:overall_architecture}
\end{figure*}

\section{Related Work}

\subsection{LiDAR-Based 3-D Object Detection}

Existing LiDAR-based methods can be broadly categorized into point-based,
voxel-based, and pillar-based approaches~\cite{qian2022survey}.
Point-based methods~\cite{qi2017pointnet,shi2019pointrcnn} directly process raw point clouds and preserve
fine-grained geometric details, but their point-wise neighborhood
construction and feature aggregation can be computationally demanding in
large-scale driving scenes.
Voxel-based methods, such as VoxelNet~\cite{zhou2018voxelnet} and SECOND~\cite{yan2018second}, convert
irregular point clouds into structured voxel representations and employ
dense or sparse convolutions for feature extraction.
Pillar-based methods, represented by PointPillars~\cite{lang2019pointpillars}, partition points
into vertical pillars for efficient BEV feature extraction.
Subsequent methods further improve LiDAR detection through hybrid point--voxel feature abstraction, center-based prediction, and fully sparse voxel processing~\cite{shi2020pvrcnn,yin2021centerpoint,chen2023voxelnext}.
SIENet mitigates distance-induced point-density imbalance by enhancing spatial structure representations, whereas HRNet progressively refines multiscale voxel features~\cite{li2022sienet,lu2024hrnet}.
BADet models local boundary correlations for object proposals through neighborhood graphs, while Objformer exploits instance-wise interaction to improve representations under point-cloud incompletion and occlusion~\cite{qian2022badet,tao2024objformer}.
Nevertheless, distance-induced sparsity continues to hinder far-range detection~\cite{li2022sienet}; incompleteness and occlusion remain challenging~\cite{lu2024hrnet,tao2024objformer}; and adverse weather further degrades LiDAR measurements and detector performance~\cite{hahner2021fog,chae2024robustfusion}.

\subsection{4-D Radar-Based 3-D Object Detection}

In recent years, 4-D radar has attracted increasing attention for 3-D
object detection.
Compared with conventional millimeter-wave radar, 4-D radar additionally
provides elevation measurements, enabling radar detections to be
represented as 3-D point clouds.
It also provides Doppler velocity and radar cross section (RCS), offering
motion and reflectivity information while maintaining robustness at long
range and under adverse weather conditions~\cite{fan2024radarsurvey,palffy2022vod}.
Since 4-D radar data share certain representation characteristics with
LiDAR point clouds, LiDAR-based detectors such as
PointPillars~\cite{lang2019pointpillars} can be directly adapted to radar
inputs.
However, such direct adaptation is generally suboptimal because of the
substantial modality differences in point density, measurement
uncertainty, and physical attributes.

Several radar-only methods therefore focus on radar-specific feature
encoding.
RPFA-Net~\cite{xu2021rpfanet} and
RadarPillars~\cite{musiat2024radarpillars} introduce attention mechanisms
to enhance sparse feature aggregation.
RadarMFNet~\cite{tan2023radarmfnet} employs velocity compensation and
multiframe aggregation, whereas DR-Net~\cite{cao2026drnet} further
exploits velocity cues through dual representations and motion-aware
augmentation.
SMURF~\cite{liu2024smurf} combines pillar and kernel-density
representations, while MAFF-Net~\cite{bi2025maffnet} suppresses radar
noise through auxiliary denoising and keypoint enhancement.
However, owing to the inherent sparsity, noise, and irregular distribution
of 4-D radar point clouds, radar-only methods generally achieve lower
detection performance than LiDAR-based methods.

\subsection{LiDAR and 4-D Radar Fusion}

Recognizing the limitations of single-modality perception, recent studies
have increasingly explored LiDAR and 4-D radar fusion to leverage their
complementary characteristics.
Existing methods explore cross-modal interaction at different
representation levels and fusion stages through diverse fusion strategies.
InterFusion~\cite{wang2022interfusion} performs early interaction between
pillarized LiDAR and radar features.
M2-Fusion~\cite{wang2023m2fusion} introduces cross-modal attention and
center-oriented multiscale aggregation to improve feature learning across
modalities.
Chae et al.~\cite{chae2024robustfusion} exploit the 3-D spatial relations
between LiDAR and radar features together with weather-conditional gating,
while DLR-Fusion~\cite{chae2025dopplerfusion} explicitly incorporates
Doppler cues through multipath interaction.
L4DR~\cite{huang2025l4dr} introduces a two-stage fusion strategy by
performing cross-modal data complementation with a 3-D fusion encoder,
followed by parallel modality-specific feature extraction and multiscale
gated fusion.
MutualForce~\cite{peng2025mutualforce} performs LiDAR and 4-D radar fusion
at both the pillar and BEV levels, while
SVEFusion~\cite{yang2026svefusion} explores voxel-level cross-modal
interaction and salient voxel enhancement.
REL~\cite{tong2026rel} employs radar features to enhance degraded LiDAR
geometry.
MoRAL extends LiDAR--4-D radar fusion to multiframe radar accumulation by
compensating motion-induced inter-frame misalignment and introducing
motion-aware gated fusion~\cite{peng2025moral}.
V2X-R explores LiDAR--4-D radar fusion in cooperative perception settings
and uses 4-D radar features to guide LiDAR feature denoising under adverse
weather conditions~\cite{huang2025v2xr}.

Despite these advances, local geometric complementation for radar pillars
without co-located LiDAR support and cross-stage radar-evidence-guided
adaptation of modality and scale responses remain underexplored.
ESAFusion propagates radar evidence across fusion stages and uses it together
with geometric consistency to guide LiDAR-to-radar complementation, while
jointly adapting modality and scale responses in BEV space.

\begin{figure*}[!t]
    \centering
    \includegraphics[
        width=\textwidth,
        trim=29bp 73bp 46bp 32bp,
        clip
    ]{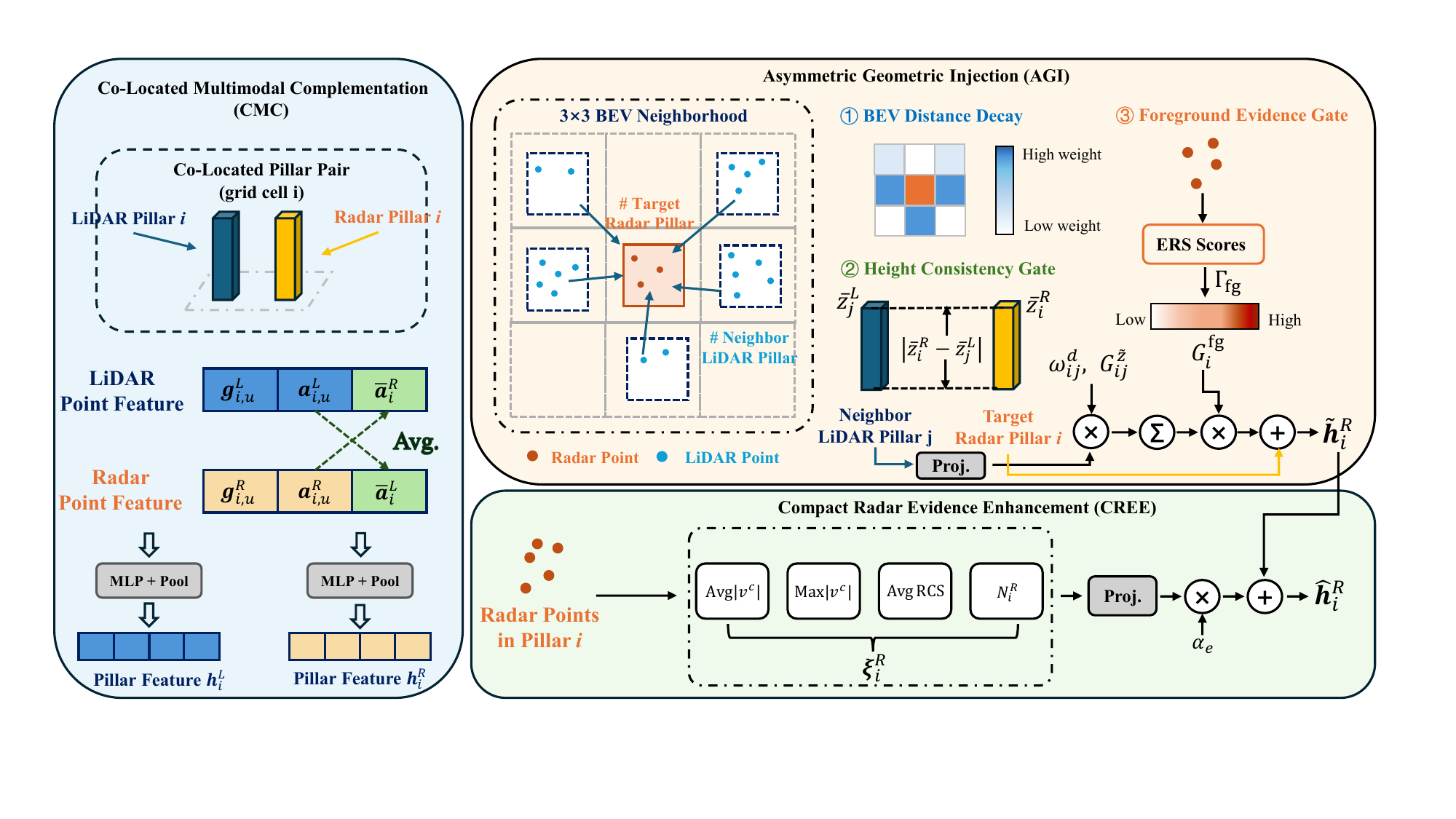}
    % Fig. 4
    \caption{Architecture of the PCE module, comprising Co-Located Multimodal Complementation (CMC), Asymmetric Geometric Injection (AGI), and Compact Radar Evidence Enhancement (CREE).}
    \label{fig:pce_architecture}
\end{figure*}

\section{Method}
\label{sec:method}

\subsection{Overview}
\label{subsec:overview}

The overall architecture of ESAFusion is illustrated in
Fig.~\ref{fig:overall_architecture}.
Given LiDAR and 4-D radar point clouds, the framework sequentially models
radar evidence at the point level, performs cross-modal complementation with
local geometric support at the pillar level, and conducts multiscale adaptive
interaction in BEV space.

First, the Evidence-Aware Radar Selection (ERS) module restructures radar
attributes using motion and observation-quality evidence, filters out
low-confidence clutter, and passes the foreground confidence scores of the
retained radar points to the subsequent pillar-level encoder.
Next, the Pillar-Level Complementary Encoder (PCE) operates within a shared
pillar grid. It performs attribute complementation across modalities within
co-located pillar pairs, incorporates local support from neighboring LiDAR
pillars based on geometric consistency, and enhances radar pillar
representations with compact radar evidence. The resulting LiDAR and radar
pillar features are scattered into a shared BEV space.
Finally, the Intra- and Inter-Scale Adaptive Fusion (ISAF) module applies
modality gating within each scale and performs region-adaptive feature
recalibration across scales. Propagated radar evidence further refines the
highest-resolution radar gate, while multiscale semantic content and local
modality observation support guide inter-scale recalibration. The resulting
fused BEV representation is fed into the 3-D detection head.

\subsection{Evidence-Aware Radar Selection (ERS)}
\label{subsec:ers}

4-D radar point clouds contain numerous unstable returns and background
clutter. Jointly modeling heterogeneous physical attributes may limit the
network's ability to distinguish target-relevant returns from unreliable
background responses. To address this issue, we propose ERS, which explicitly
restructures point-level radar attributes using motion and observation-quality
evidence and then predicts foreground confidence to filter out low-confidence
clutter. The confidence scores of retained points are passed to the subsequent
pillar-level encoder.

For the $i$-th radar point, the two types of evidence are defined as
\begin{equation}
\begin{aligned}
\boldsymbol{e}_i^m
&=
\bigl[
v_{r,i},
v_i^c,
\lvert v_{r,i} \rvert,
\lvert v_i^c \rvert,
v_{r,i}-v_i^c
\bigr], \\
\boldsymbol{e}_i^q
&=
\bigl[
\mathrm{RCS}_i,
d_i,
\lvert \Delta t_i \rvert
\bigr],
\end{aligned}
\label{eq:ers_evidence}
\end{equation}
where $v_{r,i}$ and $v_i^c$ denote the raw and ego-motion-compensated
radial velocities, respectively. $d_i$ denotes the measurement range, and
$\lvert \Delta t_i \rvert$ denotes the absolute temporal offset relative to
the reference frame. Motion evidence characterizes the radar dynamics, while
observation-quality evidence describes radar reliability. The two evidence
types are processed by independent branches to generate point-level evidence
embeddings and channel-wise gates:
\begin{equation}
\begin{aligned}
\boldsymbol{u}_i^b
&=
\operatorname{MLP}_b\bigl(\boldsymbol{e}_i^b\bigr), \\
\boldsymbol{g}_i^b
&=
\sigma\left(
\operatorname{Linear}_b\bigl(\boldsymbol{u}_i^b\bigr)
\right),
\qquad
b \in \{m,q\},
\end{aligned}
\label{eq:ers_evidence_encoding}
\end{equation}
where $b$ indexes the evidence branch, with $m$ and $q$ denoting the motion
and observation-quality branches, respectively. $\boldsymbol{u}_i^b$ is a
point-level evidence embedding with the same dimensionality as the original
radar attributes, and $\boldsymbol{g}_i^b$ denotes the corresponding channel-wise
adaptive gating weights. The original radar attributes are then restructured
through a gated residual update:
\begin{equation}
\widetilde{\boldsymbol{a}}_i^R
=
\boldsymbol{a}_i^R
+
\alpha_m\boldsymbol{g}_i^m \odot \boldsymbol{u}_i^m
+
\alpha_q\boldsymbol{g}_i^q \odot \boldsymbol{u}_i^q,
\label{eq:ers_attribute_update}
\end{equation}
where $\boldsymbol{a}_i^R$ denotes the original attributes of the $i$-th radar
point, $\alpha_m$ and $\alpha_q$ control the contributions of the two evidence
types, and $\odot$ denotes channel-wise multiplication.

Following a foreground-aware point-level selection
strategy~\cite{huang2025l4dr}, we use PointNet++~\cite{qi2017pointnetpp} with a segmentation
head as $\chi$ to predict point-wise foreground confidence
$S=\chi(\widetilde{\mathcal{P}}^{R})$, filtering out radar points with
confidence scores below a predefined threshold $\tau$ and retaining
$\mathcal{P}_{\mathrm{fg}}^R
=
\{\boldsymbol{p}_i^R \mid S_i \geq \tau\}$.
The confidence score $S_i$ of each retained point is appended as an additional
attribute channel and passed to the subsequent pillar-level encoder as explicit
foreground evidence.

\subsection{Pillar-Level Complementary Encoder (PCE)}
\label{subsec:pce}

As shown in Fig.~\ref{fig:pce_architecture}, the PCE consists of three main stages: Co-located Multimodal Complementation,
Asymmetric Geometric Injection, and Compact Radar Evidence Enhancement.

\noindent\textbf{Co-Located Multimodal Complementation (CMC).}
Following the standard pillar encoding paradigm~\cite{lang2019pointpillars},
we first partition the LiDAR and ERS-filtered radar point clouds into a shared
BEV pillar grid. Let $\mathcal{V}_i^m$ denote the set of points from modality
$m\in\{L,R\}$ assigned to grid location $i$. When both $\mathcal{V}_i^L$ and
$\mathcal{V}_i^R$ are nonempty, the LiDAR and radar pillars at this location
form a co-located pillar pair. Cross-modal attribute complementation is then
performed within each pair.

For a co-located pillar $i$, the input feature of the $u$-th point from
modality $m$ is decomposed into spatial and modality-specific attribute terms:
\begin{equation}
\boldsymbol{f}_{i,u}^{m}
=
\left[
\boldsymbol{g}_{i,u}^{m},
\boldsymbol{a}_{i,u}^{m}
\right],
\qquad
m\in\{L,R\},
\label{eq:cmc_point_feature}
\end{equation}
where $\boldsymbol{g}_{i,u}^{m}$ denotes the point-level spatial encoding
constructed from point coordinates and relative geometric offsets. These
offsets include those from the point to the pillar center and the same-modality
point-cluster center. $\boldsymbol{a}_{i,u}^{m}$ denotes
modality-specific point attributes, including LiDAR reflectance intensity,
as well as radar velocity, RCS, and the foreground confidence score generated
by ERS. For the counterpart modality $\bar{m}$, we average its within-pillar
attributes and concatenate them with the current point feature:
\begin{equation}
\overline{\boldsymbol{a}}_{i}^{\bar{m}}
=
\frac{1}{N_i^{\bar{m}}}
\sum_{v=1}^{N_i^{\bar{m}}}
\boldsymbol{a}_{i,v}^{\bar{m}},
\qquad
\widehat{\boldsymbol{f}}_{i,u}^{m}
=
\left[
\boldsymbol{g}_{i,u}^{m},
\boldsymbol{a}_{i,u}^{m},
\overline{\boldsymbol{a}}_{i}^{\bar{m}}
\right],
\label{eq:cmc_attribute_complementation}
\end{equation}
where $N_i^{\bar{m}}=\lvert\mathcal{V}_i^{\bar{m}}\rvert$ denotes the number
of counterpart-modality points within the co-located pillar. The physical
attributes of the counterpart modality are therefore shared with all points
of the current modality. Finally, the complemented point-level features are
transformed into pillar-level representations through modality-specific
within-pillar aggregation:
\begin{equation}
\boldsymbol{h}_i^{m}
=
\mathcal{A}_m
\left(
\left\{
\widehat{\boldsymbol{f}}_{i,u}^{m}
\right\}_{u=1}^{N_i^{m}}
\right),
\label{eq:cmc_pillar_aggregation}
\end{equation}
where $\mathcal{A}_m$ comprises point-level mapping and symmetric
pooling, and $\boldsymbol{h}_i^{m}$ denotes the resulting modality-specific
pillar feature.

\noindent\textbf{Asymmetric Geometric Injection (AGI).}
Because LiDAR and radar differ in sampling mechanisms, spatial resolution,
and measurement uncertainty, cross-modal observations of the same object may
occupy neighboring rather than strictly co-located BEV pillars. Valid
cross-modal geometric correspondences may still exist within local
neighborhoods even without strict co-location. Relying solely on co-located
matching would limit the spatial coverage of complementary information. To
realize local geometric complementation beyond strict co-location, we propose
Asymmetric Geometric Injection (AGI), which aggregates LiDAR
geometric and reflectance information from a local neighborhood and injects
it unidirectionally into radar pillars without co-located LiDAR support.

Specifically, for a radar pillar $i$ satisfying
$\mathcal{V}_i^R\neq\varnothing$ and $\mathcal{V}_i^L=\varnothing$, we define
the nonempty LiDAR pillars within its local $3\times3$ BEV neighborhood
$\Omega(i)$ as the candidate support set:
\begin{equation}
\mathcal{N}_i^L
=
\left\{
j\in\Omega(i)
\;\middle|\;
\mathcal{V}_j^L\neq\varnothing
\right\},
\label{eq:agi_candidate_set}
\end{equation}
where $\mathcal{N}_i^L$ denotes the set of LiDAR pillars that can provide
neighborhood information for the target radar pillar $i$.

Since the candidate LiDAR pillars differ in their spatial proximity to the
target radar pillar, we assign local decay weights according to their BEV
distances:
\begin{equation}
\omega_{ij}^{d}
=
\frac{1}{
1+
\left\|
\boldsymbol{c}_i^R-\boldsymbol{c}_j^L
\right\|_2^2
},
\label{eq:agi_distance_weight}
\end{equation}
where $\boldsymbol{c}_i^R$ and $\boldsymbol{c}_j^L$ denote the BEV center
coordinates of target radar pillar $i$ and candidate LiDAR pillar $j$,
respectively. A closer candidate LiDAR pillar therefore contributes more
strongly to the local support.

However, BEV proximity does not guarantee consistent geometric correspondence
between the two modalities in 3-D space. We further exploit the elevation
measurements provided by 4-D radar, compare the mean heights of the target radar
pillar and each candidate LiDAR pillar, and construct a height-consistency gate:
\begin{equation}
G_{ij}^{\widetilde{z}}
=
\max
\left\{
G_{\min},
\exp
\left(
-
\frac{
\left(
\overline{z}_i^R-\overline{z}_j^L
\right)^2
}{
2\sigma_z^2
}
\right)
\right\},
\label{eq:agi_height_gate}
\end{equation}
where $\overline{z}_i^R$ and $\overline{z}_j^L$ denote the mean point heights
within target radar pillar $i$ and candidate LiDAR pillar $j$, respectively.
$\sigma_z$ controls the sensitivity of the gate to height differences, and
$G_{\min}$ denotes the minimum retained weight.

To prevent low-confidence radar pillars from indiscriminately receiving
neighboring LiDAR information, we introduce the pillar-level foreground
evidence aggregation function $\Gamma_{\mathrm{fg}}(\cdot)$ to convert the
point-level foreground scores retained by ERS into a continuous gate for each
radar pillar:
\begin{equation}
G_i^{\mathrm{fg}}
=
\Gamma_{\mathrm{fg}}
\left(
\left\{
S_{i,u}
\right\}_{u=1}^{N_i^R}
\right).
\label{eq:agi_foreground_gate}
\end{equation}
A higher overall foreground confidence within the pillar produces a larger
$G_i^{\mathrm{fg}}$, allowing stronger support from neighboring LiDAR pillars.
After weighting by BEV distance, height consistency, and foreground evidence,
the candidate LiDAR pillar features are projected into a representation space
compatible with radar pillar features.
They are then aggregated as local support for the target radar pillar:
\begin{equation}
\boldsymbol{s}_i^{L\rightarrow R}
=
G_i^{\mathrm{fg}}
\sum_{j\in\mathcal{N}_i^L}
\omega_{ij}^{d}
G_{ij}^{\widetilde{z}}
\psi_L
\left(
\boldsymbol{h}_j^L
\right),
\label{eq:agi_lidar_support}
\end{equation}
where $\psi_L(\cdot)$ denotes a lightweight feature projection layer, while
$\boldsymbol{s}_i^{L\rightarrow R}$ denotes the neighboring LiDAR support
aggregated for target radar pillar $i$.

The aggregated LiDAR support is subsequently incorporated into the
non-co-located target radar pillar representation through a controlled update:
\begin{equation}
\widetilde{\boldsymbol{h}}_i^R
=
\boldsymbol{h}_i^R
+
\beta
\boldsymbol{s}_i^{L\rightarrow R},
\label{eq:agi_radar_update}
\end{equation}
where $\beta$ controls the fusion strength of the local support, and
$\widetilde{\boldsymbol{h}}_i^R$ denotes the enhanced radar pillar feature.
Since LiDAR provides denser and more stable local geometric structures,
whereas radar observations are relatively sparse and susceptible to clutter,
AGI performs only LiDAR-to-radar geometric injection.

\begin{figure*}[!t]
    \centering
    \includegraphics[
        width=\textwidth,
        trim=78bp 32bp 71bp 38bp,
        clip
    ]{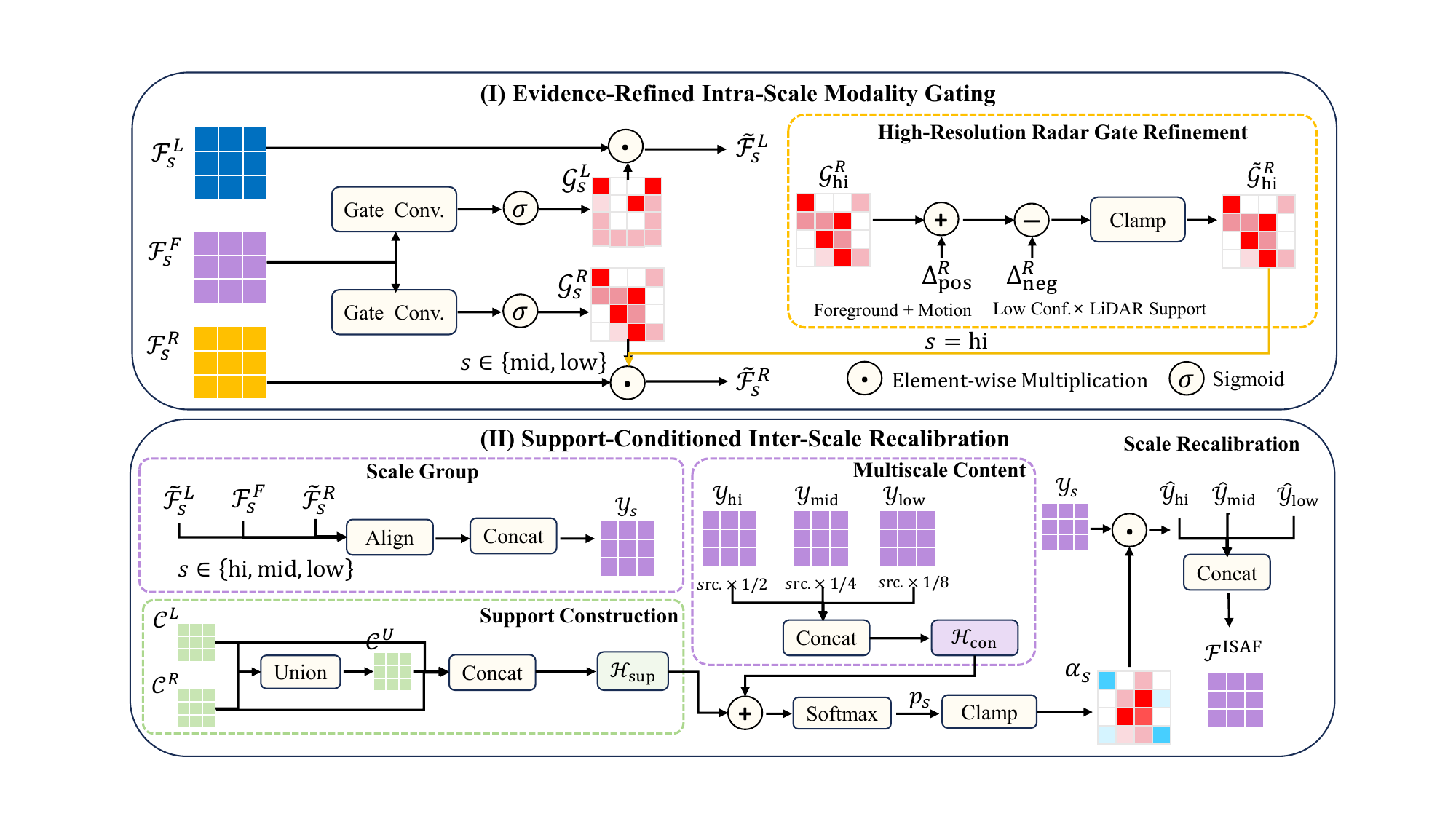}
    % Fig. 5
    \caption{Architecture of the ISAF module, comprising Evidence-Refined Intra-Scale Modality Gating and Support-Conditioned Inter-Scale Recalibration.}
    \label{fig:isaf_architecture}
\end{figure*}

\noindent\textbf{Compact Radar Evidence Enhancement (CREE).}
Although radar physical attributes are included in the point-level inputs,
they become entangled in high-dimensional features after nonlinear
point-to-pillar mapping, aggregation, and cross-modal interaction, making
their physical and statistical information difficult to exploit directly.
We therefore summarize the compensated radial velocities, RCS, and the number
of valid radar points within each pillar to construct compact pillar-level
radar evidence, which is then used to explicitly enhance the radar pillar
features.

For radar pillar $i$, the compact radar evidence is defined as:
\begin{equation}
\boldsymbol{\xi}_i^R
=
\left[
\operatorname{Avg}_{u}
\left\lvert v_{i,u}^{c} \right\rvert,
\operatorname{Max}_{u}
\left\lvert v_{i,u}^{c} \right\rvert,
\operatorname{Avg}_{u}
\mathrm{RCS}_{i,u},
N_i^R
\right],
\label{eq:cree_compact_evidence}
\end{equation}
where the mean and maximum absolute compensated radial velocities characterize
the overall motion level and strongest local motion response within the pillar,
respectively. The mean RCS represents the overall return response, while
$N_i^R$ reflects the degree of observational support for the radar pillar.
The compact evidence is then projected into the pillar feature space to
enhance the radar pillar feature:
\begin{equation}
\widehat{\boldsymbol{h}}_i^R
=
\widetilde{\boldsymbol{h}}_i^R
+
\alpha_e
\phi_e
\left(
\boldsymbol{\xi}_i^R
\right),
\label{eq:cree_evidence_enhancement}
\end{equation}
where $\phi_e(\cdot)$ maps the compact radar evidence into the pillar feature
space, and $\alpha_e$ controls the evidence-enhancement strength. The resulting
$\widehat{\boldsymbol{h}}_i^R$ denotes the final radar pillar feature output
by PCE.

In this way, building upon co-located attribute complementation, PCE extends
cross-modal support to local neighborhoods by weighting neighboring LiDAR
support according to BEV proximity, height consistency, and radar foreground
evidence, and further explicitly strengthens radar-specific pillar-level
physical evidence.

\subsection{Intra- and Inter-Scale Adaptive Fusion (ISAF)}
\label{subsec:isaf}

As shown in Fig.~\ref{fig:isaf_architecture}, the PCE-encoded LiDAR and
radar pillar features are scattered into a shared BEV space. ISAF implements
multiscale adaptive interaction by first regulating modality-specific responses
within each scale and then recalibrating scale responses according to multiscale
semantic content and local observation support. Specifically, it modulates
LiDAR and radar responses within each scale using fused context and refines the
highest-resolution radar gate with radar foreground confidence and motion
evidence propagated from the front end. It then performs bounded,
region-adaptive recalibration of the spatially aligned scale groups, ultimately
producing the fused BEV feature.

\noindent\textbf{Evidence-Refined Intra-Scale Modality Gating.}
After being encoded by PCE and scattered into the BEV space, the LiDAR and
radar features $\mathcal{F}^{L}$ and $\mathcal{F}^{R}$ are fused to construct
a joint BEV context feature $\mathcal{F}^{F}$:
\begin{equation}
\mathcal{F}^{F}
=
\phi
\left(
\mathcal{F}^{L},
\mathcal{F}^{R}
\right),
\label{eq:isaf_fused_context}
\end{equation}
where $\phi(\cdot)$ denotes concatenation along the channel dimension. We then
perform hierarchical multiscale encoding along three parallel LiDAR, radar,
and fused branches:
\begin{equation}
\mathcal{F}_s^m
=
\kappa
\left(
\mathcal{F}_{\operatorname{pre}(s)}^m
\right),
m\in\{L,R,F\},
s\in\{\mathrm{hi},\mathrm{mid},\mathrm{low}\},
\label{eq:isaf_multiscale_encoding}
\end{equation}
where $s$ denotes the feature scale, and $\operatorname{pre}(s)$ denotes its
preceding scale. $m$ indicates different modality, and $\kappa$ represents a
convolutional block with batch normalization and ReLU activation.

Following the multiscale gated fusion strategy in L4DR~\cite{huang2025l4dr}, we generate
separate LiDAR and radar gates from the fused context at each scale:
\begin{equation}
\mathcal{G}_s^m
=
\sigma
\left(
W_s^{F\rightarrow m}
\ast
\mathcal{F}_s^F
\right),
\qquad
m\in\{L,R\},
\label{eq:isaf_base_gate}
\end{equation}
where $\mathcal{G}_s^m$ denotes the corresponding modality gating map,
$W_s^{F\rightarrow m}$ is a learnable convolution kernel that generates the
gate for modality $m$ from the fused context at scale $s$, $\ast$ denotes
convolution, and $\sigma(\cdot)$ is the Sigmoid activation function.

However, base gates generated solely from fused semantic context cannot
explicitly exploit the foreground confidence and motion information in radar
observations. Since such evidence exhibits strong local spatial correspondence
and is better preserved in high-resolution BEV features, we construct separate
positive and negative evidence refinement terms only at the highest-resolution
scale:
\begin{equation}
\Delta_{\mathrm{pos}}^{R}
=
E_{\mathrm{conf}}^{R,+}
+
\beta_{\mathrm{mot}}
E_{\mathrm{mot}}^{R},
\quad
\Delta_{\mathrm{neg}}^{R}
=
S_{\mathrm{hi}}^{L}
\odot
E_{\mathrm{conf}}^{R,-},
\label{eq:isaf_evidence_refinement}
\end{equation}
where $E_{\mathrm{conf}}^{R,+}$ denotes positive confidence
evidence constructed from high-confidence radar foreground scores, and
$E_{\mathrm{mot}}^R$ denotes radar motion evidence constrained
to high-confidence regions. $\beta_{\mathrm{mot}}$ balances their relative
contributions. Similarly, $E_{\mathrm{conf}}^{R,-}$ denotes
negative confidence evidence constructed from low-confidence radar foreground
scores, while $S_{\mathrm{hi}}^L$ represents local observation
support predicted from the highest-resolution LiDAR features and serves as a
cross-modal validation condition for negative suppression. The positive
refinement combines radar foreground confidence with motion evidence to
enhance reliable radar responses in potential target regions, while the
negative refinement suppresses responses only where low radar confidence
coincides with local LiDAR support, thereby avoiding excessive suppression
of potentially valid sparse radar observations without co-located LiDAR
support. Then we use the above evidence to refine the highest-resolution
radar gate:
\begin{equation}
\widetilde{\mathcal{G}}_{\mathrm{hi}}^{R}
=
\operatorname{Clamp}
\left(
\mathcal{G}_{\mathrm{hi}}^{R}
+
\lambda_{+}
\Delta_{\mathrm{pos}}^{R}
-
\lambda_{-}
\Delta_{\mathrm{neg}}^{R},
0,1
\right),
\label{eq:isaf_refined_radar_gate}
\end{equation}
where $\lambda_{+}$ and $\lambda_{-}$ control the overall strengths of
positive enhancement and negative suppression, respectively, and
$\operatorname{Clamp}(\cdot,0,1)$ bounds the refined gate within $[0,1]$.
$\widetilde{\mathcal{G}}_{\mathrm{hi}}^R$ denotes the refined
highest-resolution radar gate. At the mid- and low-resolution scales, the
base radar gates remain unchanged.

Finally, these gates are applied to the corresponding modality features via
element-wise multiplication to obtain the intra-scale modulated features:
\begin{equation}
\widetilde{\mathcal{F}}_s^L
=
\mathcal{G}_s^L
\odot
\mathcal{F}_s^L,
\qquad
\widetilde{\mathcal{F}}_s^R
=
\widetilde{\mathcal{G}}_s^R
\odot
\mathcal{F}_s^R.
\label{eq:isaf_intrascale_features}
\end{equation}
In this way, the intra-scale gating adaptively modulates LiDAR and radar
feature responses at each scale based on the fused context.

\begin{table*}[!t]
\centering
\caption{Comparative AP (\%) results on the VoD validation set. The best results are
bold, and the second best are underlined. R, C, and L denote 4-D radar, camera,
and LiDAR, respectively}
\label{tab:vod_main_results}
\small
\renewcommand{\arraystretch}{1.05}
\begin{tabular*}{\textwidth}{@{\hspace{6pt}\extracolsep{\fill}} l c *{8}{c} }
\toprule
\multirow{2}{*}{Methods} & \multirow{2}{*}{Modality}
& \multicolumn{4}{c}{Entire Annotated Area (EAA)}
& \multicolumn{4}{c}{Driving Corridor (RoI)} \\
\cmidrule(lr){3-6}\cmidrule(l){7-10}
& & Car & Ped. & Cyc. & mAP & Car & Ped. & Cyc. & mAP \\
\midrule
PointPillars (CVPR'19)~\cite{lang2019pointpillars}        & R   & 38.89 & 32.07 & 65.02 & 45.32 & 71.07 & 42.61 & 86.61 & 66.77 \\
SMURF (T-IV'24)~\cite{liu2024smurf}              & R   & 42.31 & 39.09 & 71.50 & 50.97 & 71.74 & 50.54 & 86.87 & 69.72 \\
4DRadDet (ICRA'25)~\cite{weng2025raddet}           & R   & 42.03 & 40.70 & 71.61 & 51.44 & 72.12 & 51.18 & 87.95 & 70.42 \\
MAFF-Net (RA-L'25)~\cite{bi2025maffnet}           & R   & 42.33 & 46.75 & 74.72 & 54.59 & 72.28 & 57.81 & 87.40 & 72.50 \\
RadarGaussianDet3D (RA-L'26)~\cite{xiong2026radargaussiandet3d} & R   & 40.70 & 42.40 & 73.00 & 52.03 & 71.20 & 51.70 & 89.00 & 70.63 \\
\midrule
LXL (T-IV'24)~\cite{xiong2024lxl}                & C+R & 42.33 & 49.48 & 77.12 & 56.31 & 72.18 & 58.30 & 88.31 & 72.93 \\
UniBEVFusion (ICRA'25)~\cite{zhao2025unibevfusion}       & C+R & 42.22 & 47.11 & 72.94 & 54.09 & 72.10 & 57.71 & 93.29 & 74.37 \\
SGDet3D (RA-L'25)~\cite{bai2025sgdet3d}            & C+R & 53.16 & 49.98 & 76.11 & 59.75 & 81.13 & 60.91 & 90.22 & 77.42 \\
ELAFusion (RA-L'26)~\cite{hu2026elafusion}          & C+R & 57.73 & 58.59 & 78.79 & 65.04 & 89.48 & 68.66 & 95.40 & 84.51 \\
RPGFusion (CVPR'26)~\cite{qiu2026rpgfusion}          & C+R & 67.37 & 59.94 & 80.62 & 69.31 & \textbf{94.42} & 71.63 & 92.55 & 86.20 \\
\midrule
PointPillars (CVPR'19)~\cite{lang2019pointpillars}       & L   & 65.55 & 55.71 & 72.96 & 64.74 & 81.10 & 67.92 & 88.96 & 79.33 \\
Voxel Mamba (NeurIPS'24)~\cite{zhang2024voxelmamba}     & L   & 67.28 & 65.34 & 74.38 & 69.00 & 90.67 & 78.14 & 86.40 & 85.07 \\
\midrule
InterFusion (IROS'22)~\cite{wang2022interfusion}        & L+R & 66.50 & 64.50 & 78.50 & 69.83 & 90.70 & 72.00 & 88.70 & 83.80 \\
RLNet (ECCV'24)~\cite{xu2025rlnet}              & L+R & 70.88 & \textbf{69.43} & 78.12 & 72.81 & 90.82 & 78.71 & 91.67 & 87.08 \\
CM-FA (ICRA'24)~\cite{deng2024cmfa}              & L+R & \textbf{77.52} & 67.99 & 75.97 & 73.83 & 90.91 & \textbf{80.78} & 87.80 & 86.50 \\
L4DR (AAAI'25)~\cite{huang2025l4dr}               & L+R & 69.10 & 66.20 & 82.80 & 72.70 & 90.80 & 76.10 & 95.50 & 87.47 \\
SVEFusion (PR'26)~\cite{yang2026svefusion}            & L+R & \underline{71.83} & 67.85 & \underline{83.97} & \underline{74.55} & 90.91 & \underline{79.37} & \underline{96.31} & \underline{88.86} \\
ESAFusion (ours)                    & L+R & 70.43 & \underline{69.24} & \textbf{84.12} & \textbf{74.60} & \underline{90.92} & 78.59 & \textbf{97.16} & \textbf{88.89} \\
\bottomrule
\end{tabular*}
\end{table*}

\noindent\textbf{Support-Conditioned Inter-Scale Recalibration.}
Multiscale features play complementary roles: high-resolution features
preserve fine geometric details and localized radar motion responses, while
mid- and low-resolution features capture broader spatial context and
object-level structures. Direct multiscale concatenation lacks region-adaptive
scale modulation, making it difficult to adjust the relative contribution of
each scale according to the representational demands of different BEV regions.

To address this limitation, we perform bounded, region-adaptive recalibration
of the spatially aligned scale groups conditioned on multiscale semantics and
local observation support from both modalities. After intra-scale gating, the
fused, LiDAR and radar branch features at each scale are first aligned to a
common BEV resolution and then concatenated along the channel dimension to
form the corresponding scale group:
\begin{equation}
\mathcal{Y}_s
=
\operatorname{Concat}
\left[
\mathcal{U}_s^F
\left(
\mathcal{F}_s^F
\right),
\mathcal{U}_s^L
\left(
\widetilde{\mathcal{F}}_s^L
\right),
\mathcal{U}_s^R
\left(
\widetilde{\mathcal{F}}_s^R
\right)
\right],
\label{eq:isaf_scale_group}
\end{equation}
where $\mathcal{U}_s^m(\cdot)$ denotes spatial alignment for branch $m$, and
$\mathcal{Y}_s$ is the resulting scale group. To characterize the local
observation status of each modality across BEV regions, we first construct
binary occupancy maps from the LiDAR and radar BEV features and then apply
local average pooling to obtain modality-specific observation support:
\begin{equation}
\begin{gathered}
\mathcal{O}^m
=
\mathbb{I}
\left[
\max_c
\left|
\mathcal{F}_c^m
\right|>0
\right],
\quad
\mathcal{C}^m
=
\mathcal{A}_K
\left(
\mathcal{O}^m
\right), \\
\mathcal{C}^U
=
\mathbb{I}
\left[
\max
\left(
\mathcal{C}^L,
\mathcal{C}^R
\right)>0
\right],
\quad
\mathcal{C}
=
\operatorname{Concat}
\left[
\mathcal{C}^L,
\mathcal{C}^R,
\mathcal{C}^U
\right],
\end{gathered}
\label{eq:isaf_observation_support}
\end{equation}
where $\mathcal{O}^m$ denotes the BEV occupancy map of modality $m$, and
$\mathcal{A}_K$ denotes local average pooling. $\mathcal{C}^L$ and
$\mathcal{C}^R$ characterize the local observation coverage of LiDAR and
radar, respectively, while $\mathcal{C}^U$ indicates whether the neighborhood
contains observations from at least one modality. Together, the three maps
form the local modality observation support $\mathcal{C}$.

Multiscale semantic content characterizes the scale requirements of different
BEV regions, while local modality observation support provides the
corresponding observation conditions. Together, they predict spatially
varying scale responses:
\begin{equation}
p_s
=
\left[
\operatorname{Softmax}_{s}
\left(
\mathcal{H}_{\mathrm{con}}
\left(
\operatorname{Concat}_{k\in\mathcal{S}}
\mathcal{Y}_k
\right)
+
\mathcal{H}_{\mathrm{sup}}
\left(
\mathcal{C}
\right)
\right)
\right]_s,
\label{eq:isaf_scale_response}
\end{equation}
where $\mathcal{H}_{\mathrm{con}}$ generates the scale prediction term from
multiscale semantic content, while $\mathcal{H}_{\mathrm{sup}}$ generates the
support-conditioning term from local modality observation support. $p_s$
denotes the normalized response of scale $s$ at each BEV location. We further
convert it into a bounded recalibration coefficient centered at unity:
\begin{equation}
\alpha_s
=
\operatorname{Clamp}
\left(
1+\rho(3p_s-1),
\alpha_{\min},
\alpha_{\max}
\right),
\label{eq:isaf_recalibration_coefficient}
\end{equation}
where $\rho$ controls the recalibration magnitude, while $\alpha_{\min}$ and
$\alpha_{\max}$ bound the adjustment range. Finally, we use $\alpha_s$ to
recalibrate the entire scale group $\mathcal{Y}_s$ and concatenate the
resulting groups across all scales to obtain the enhanced BEV feature:
\begin{equation}
\widehat{\mathcal{Y}}_s
=
\alpha_s
\odot
\mathcal{Y}_s,
\qquad
\mathcal{F}^{\mathrm{ISAF}}
=
\operatorname{Concat}_{s\in\mathcal{S}}
\left(
\widehat{\mathcal{Y}}_s
\right).
\label{eq:isaf_output}
\end{equation}
In this way, the mechanism preserves complementary multiscale information
while adaptively adjusting the relative contributions of different scales
according to region-specific semantic content and local modality observation
support.

\begin{table*}[!t]
\centering
\caption{Comparative 3-D and BEV AP (\%) results for the Car class on the Astyx HiRes2019 validation set at an IoU threshold of 0.5. The best results are bold, and the second best are underlined. R and L denote 4-D radar and LiDAR, respectively}
\label{tab:astyx_results}
\small
\renewcommand{\arraystretch}{1.05}
\begin{tabular*}{\textwidth}{@{\hspace{6pt}\extracolsep{\fill}} l c *{8}{c} }
\toprule
\multirow{2}{*}{Methods} & \multirow{2}{*}{Modality}
& \multicolumn{4}{c}{3-D AP}
& \multicolumn{4}{c}{BEV AP} \\
\cmidrule(lr){3-6}\cmidrule(l){7-10}
& & Easy & Moderate & Hard & Average & Easy & Moderate & Hard & Average \\
\midrule
PointPillars (CVPR'19)~\cite{lang2019pointpillars} & R & 21.58 & 15.21 & 14.15 & 16.98 & 39.52 & 27.85 & 25.75 & 31.04 \\
PV-RCNN (CVPR'20)~\cite{shi2020pvrcnn} & R & 18.37 & 13.04 & 11.95 & 14.45 & 38.82 & 27.37 & 25.54 & 30.58 \\
PV-RCNN++ (IJCV'23)~\cite{shi2023pvrcnnpp} & R & 18.20 & 13.05 & 12.04 & 14.43 & 31.73 & 23.65 & 22.24 & 25.87 \\
Voxel Mamba (NeurIPS'24)~\cite{zhang2024voxelmamba} & R & 20.63 & 14.70 & 13.89 & 16.41 & 36.27 & 27.66 & 25.87 & 29.93 \\
\midrule
PointPillars (CVPR'19)~\cite{lang2019pointpillars} & L & 47.30 & 35.05 & 32.75 & 38.37 & 52.76 & 39.79 & 37.40 & 43.32 \\
HEDNet (NeurIPS'23)~\cite{zhang2023hednet} & L & 44.83 & 34.26 & 31.98 & 37.02 & 48.10 & 37.49 & 35.24 & 40.28 \\
SAFDNet (CVPR'24)~\cite{zhang2024safdnet} & L & 49.23 & 40.54 & 38.43 & 42.73 & 53.36 & 43.89 & 41.72 & 46.32 \\
Voxel Mamba (NeurIPS'24)~\cite{zhang2024voxelmamba} & L & 51.59 & 41.09 & 38.86 & 43.85 & 56.12 & 44.74 & 42.27 & 47.71 \\
\midrule
InterFusion (IROS'22)~\cite{wang2022interfusion} & L+R & 53.73 & 43.23 & 41.35 & 46.10 & 60.14 & 48.98 & 46.83 & 51.98 \\
L4DR (AAAI'25)~\cite{huang2025l4dr} & L+R & 54.62 & 42.25 & 40.25 & 45.71 & 60.71 & 49.02 & 46.96 & 52.23 \\
SVEFusion (PR'26)~\cite{yang2026svefusion} & L+R & \underline{56.47} & \underline{44.13} & \underline{41.80} & \underline{47.47} & \underline{62.24} & \underline{49.68} & \underline{47.34} & \underline{53.09} \\
ESAFusion (ours) & L+R & \textbf{57.42} & \textbf{48.94} & \textbf{47.13} & \textbf{51.16} & \textbf{63.46} & \textbf{51.16} & \textbf{50.75} & \textbf{55.12} \\
\bottomrule
\end{tabular*}
\end{table*}

\begin{figure*}[!t]
    \centering
    \begin{minipage}[c][0.3079\textwidth][c]{\textwidth}
        \centering
        \includegraphics[
            width=\textwidth,
            height=0.3079\textwidth,
            keepaspectratio
        ]{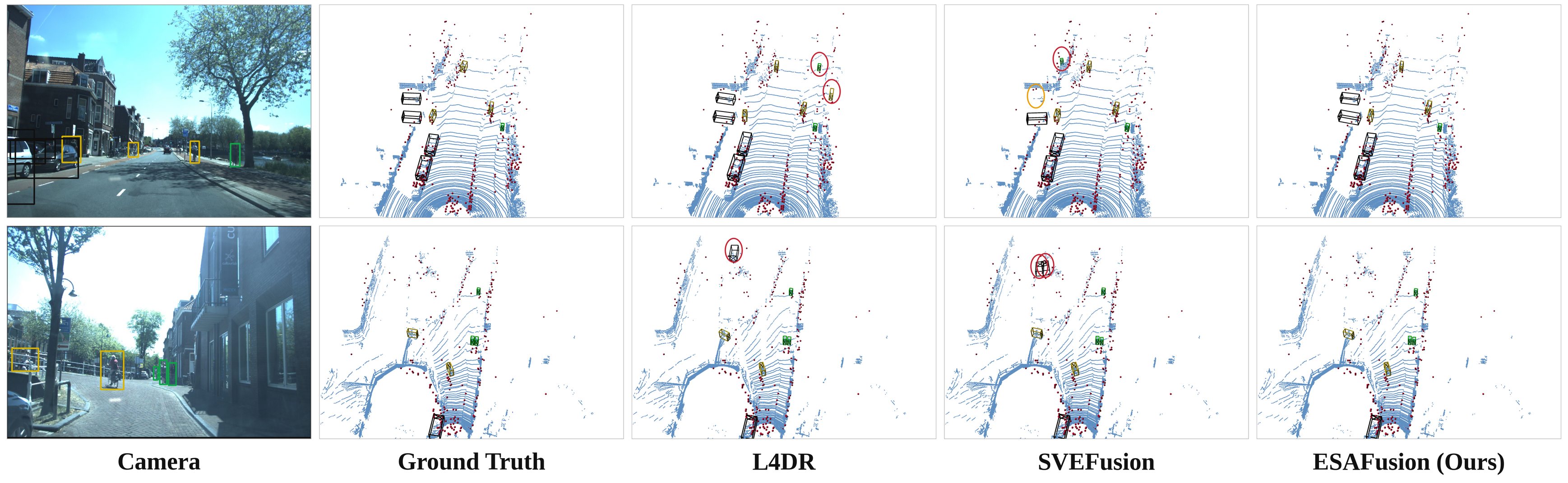}
    \end{minipage}
    \caption{Qualitative comparison of L4DR, SVEFusion, and the proposed ESAFusion on the VoD validation set. Green, yellow, and black boxes denote pedestrians, cyclists, and cars, respectively. Orange and red circles indicate false negatives and false positives, respectively.}
    \label{fig:qualitative_results}
\end{figure*}

\begin{figure}[!t]
    \centering
    \includegraphics[width=\columnwidth]{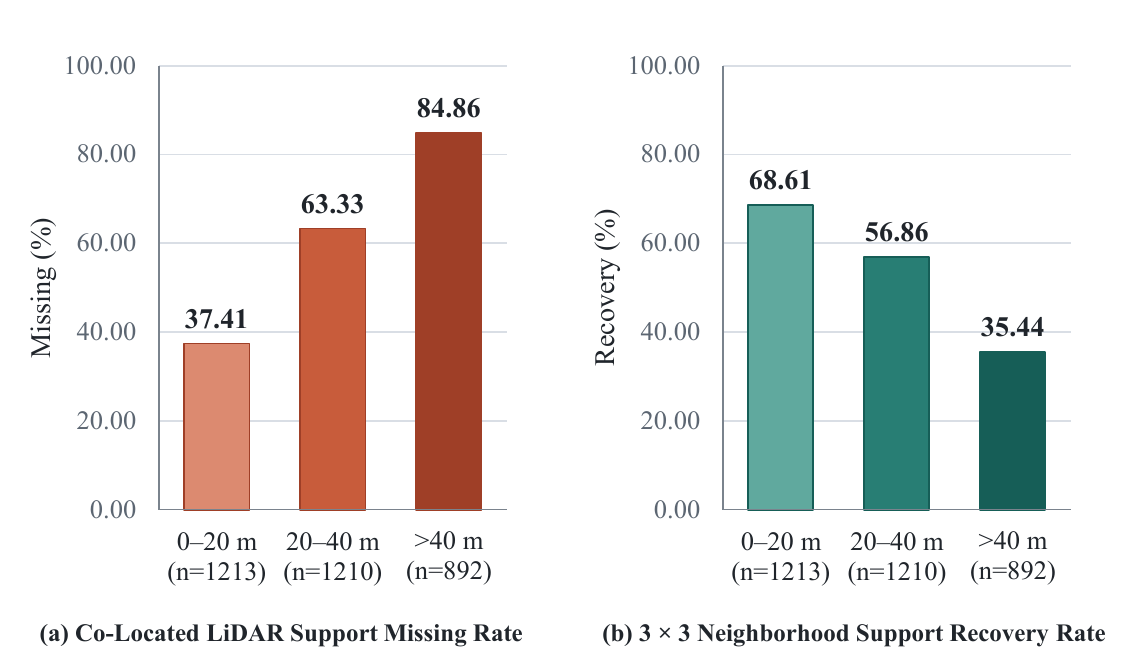}
    \caption{Distance-wise cross-modal support availability on the VoD
    validation set. (a) Percentage of radar pillars without LiDAR observations
    in the same BEV cell. (b) Among these missing-support pillars, the percentage
    having at least one LiDAR-supported cell within the local $3\times3$
    neighborhood. Here, $n$ denotes the number of valid validation frames in
    each distance interval.}
    \label{fig:distance_support}
\end{figure}

\section{Experiments}
\label{sec:experiments}

\subsection{Experimental Setup}

\noindent\textbf{Datasets and Evaluation Metrics}.
We evaluate ESAFusion on two publicly available datasets, namely the
View-of-Delft (VoD) dataset~\cite{palffy2022vod} and the Astyx
HiRes2019 dataset~\cite{meyer2019astyx}.

The VoD dataset is used as the primary benchmark for comprehensive evaluation, we follow the official split with 5,139 training frames and 1,296 validation frames.
The input consists of 5-frame accumulated 4-D radar point clouds and
single-frame LiDAR point clouds.
Following the official VoD protocol, we report 3-D average precision (AP) and mean average precision (mAP) for Car,
Pedestrian, and Cyclist in both the Entire Annotated Area (EAA) and the
Driving corridor (RoI), with intersection over union (IoU) thresholds of 0.5 for Car and 0.25 for
Pedestrian and Cyclist.

The Astyx HiRes2019 dataset provides synchronized high-resolution
4-D radar, 16-line LiDAR, and camera data, together with 3-D
annotations for seven object categories~\cite{meyer2019astyx}.
Following prior studies~\cite{wang2022interfusion,wang2023m2fusion},
we split the 546 frames into 410 training frames and 136 validation
frames. We evaluate the Car class using the KITTI-style 3-D and BEV
AP metrics at an IoU threshold of 0.5.

For robustness evaluation, we use simulated foggy VoD data following prior
settings~\cite{hahner2021fog,huang2025l4dr}.
The fog level is set to $L \in \{0, 1, 2, 3, 4\}$, with density coefficients
$\alpha = [0.00, 0.03, 0.06, 0.10, 0.20]$.
Only LiDAR point clouds are degraded, while 4-D radar data remain unchanged
to simulate various weather conditions.
We further adopt KITTI-style Easy, Moderate, and Hard difficulty splits
under different fog intensities.

\noindent\textbf{Implementation Details}.
ESAFusion is trained for 100 epochs on a single NVIDIA RTX 5090 GPU with
a batch size 8, using AdamOneCycle with a peak learning rate of 0.003 and
weight decay of 0.01.
We set the AGI neighborhood to $3\times3$, the height-gating bandwidth
$\sigma_z$ to $0.75\,\mathrm{m}$, the LiDAR-to-radar residual injection
coefficient $\beta$ to 0.02, and the ISAF scale-recalibration strength
$\rho$ to 0.05.
Other parameter settings refer to the default official configuration in
the OpenPCDet~\cite{openpcdet2020}. For efficiency evaluation, inference latency is measured on a single NVIDIA RTX 5090 GPU with a batch size of 1 after 50 warm-up batches.

\subsection{Performance Comparisons}

\noindent\textbf{Results on the VoD Dataset}.
We compare ESAFusion with existing single- and multimodal 3-D object
detection approaches in Table~\ref{tab:vod_main_results}.
Radar-only and radar--camera fusion methods generally deliver limited
detection performance, partly due to the sparsity of radar observations.
In contrast, LiDAR--radar methods achieve stronger overall performance
by combining accurate 3-D geometric information from LiDAR point clouds
with motion cues from radar.
Our ESAFusion achieves the best mAP of 74.60\% and 88.89\% in the Entire
Area and Driving Corridor, outperforming all compared methods.
Notably, for Cyclist, our method achieves the best detections with
84.12\% AP and 97.16\% AP, respectively.
Cyclists are typically represented by sparse point observations and
exhibit distinct motion patterns, making local geometric and motion cues
particularly important.
ESAFusion enhances radar pillar features with compact radar evidence and
geometrically consistent local LiDAR support, while adaptively balancing
modality- and scale-specific feature responses in the BEV backend.
This combination better preserves the local discriminative information of
dynamic targets during cross-modal fusion.
Additionally, our model delivers a real-time inference speed of
19.23~FPS.

Fig.~\ref{fig:qualitative_results} shows the qualitative results of
L4DR~\cite{huang2025l4dr}, SVEFusion~\cite{yang2026svefusion}, and our
ESAFusion.
Compared to the other two methods, ESAFusion produces fewer missed
detections and false positives.

\noindent\textbf{Results on the Astyx Dataset}.
Table~\ref{tab:astyx_results} compares ESAFusion with existing radar-only, LiDAR-only, and LiDAR--radar fusion methods on the Astyx dataset.
ESAFusion achieves the best results across all difficulty levels for both 3-D and BEV detection.
In particular, it surpasses SVEFusion~\cite{yang2026svefusion} by 3.69\% and 2.03\% in average 3-D and BEV AP, respectively.
These results further validate the effectiveness of the proposed evidence-aware and scale-adaptive fusion framework on the Astyx dataset.

\noindent\textbf{Distance-Wise Analysis.}
To examine how object range affects the performance of LiDAR--4-D radar
fusion, we conduct a distance-wise analysis on the VoD validation set,
considering both cross-modal support availability and detection performance.
Fig.~\ref{fig:distance_support} summarizes the cross-modal support statistics
across different distance intervals. The proportion of radar pillars lacking
LiDAR observations in the same BEV cell increases with range, indicating that
strict co-location becomes increasingly difficult to satisfy at long
distances. Nevertheless, some of these radar pillars still have at least one
LiDAR-supported cell within the $3 \times 3$ neighborhood. These observations
provide empirical support for the design rationale of the local geometric
complementation adopted in PCE.

\begin{table}[!t]
\centering
\caption{Distance-wise AP (\%) results on the VoD validation set. The best and
second-best results within each distance interval are highlighted in bold and
underlined, respectively}
\label{tab:distance_results}
\footnotesize
\renewcommand{\arraystretch}{1.08}
\setlength{\tabcolsep}{2.5pt}
\begin{tabular*}{\columnwidth}{@{\extracolsep{\fill}} l l c c c c}
\toprule
Distance & Method & Car AP & Ped. AP & Cyc. AP & mAP \\
\midrule
\multirow{3}{*}{0--20 m}
& L4DR~\cite{huang2025l4dr}      & 81.44 & 75.31 & 83.99 & 80.25 \\
& SVEFusion~\cite{yang2026svefusion} & \textbf{81.73} & \textbf{77.61} & \underline{85.53} & \textbf{81.62} \\
& ESAFusion & \underline{81.68} & \underline{76.62} & \textbf{86.04} & \underline{81.45} \\
\midrule
\multirow{3}{*}{20--40 m}
& L4DR~\cite{huang2025l4dr}      & 71.21 & 57.56 & \underline{88.49} & 72.42 \\
& SVEFusion~\cite{yang2026svefusion} & \textbf{72.54} & \underline{65.56} & 87.87 & \underline{75.32} \\
& ESAFusion & \underline{72.10} & \textbf{67.56} & \textbf{88.67} & \textbf{76.11} \\
\midrule
\multirow{3}{*}{$>40$ m}
& L4DR~\cite{huang2025l4dr}      & 54.41 & \underline{49.16} & \underline{58.45} & \underline{54.01} \\
& SVEFusion~\cite{yang2026svefusion} & \textbf{56.72} & 49.11 & 51.30 & 52.38 \\
& ESAFusion & \underline{56.34} & \textbf{53.88} & \textbf{61.93} & \textbf{57.38} \\
\bottomrule
\end{tabular*}
\end{table}

Table~\ref{tab:distance_results} reports the corresponding distance-wise
detection results. The overall mAP decreases with range for all methods.
ESAFusion remains comparable to SVEFusion at 0--20 m, while achieving gains of
0.79 and 5.00 mAP points at 20--40 m and beyond 40 m, respectively. In the
range beyond 40 m, ESAFusion also outperforms L4DR by 3.37 mAP points.
Relative to SVEFusion, its long-range Pedestrian and Cyclist APs increase by
4.77 and 10.63 points, respectively, while Car AP remains comparable to that
of SVEFusion. These results indicate that ESAFusion is particularly beneficial
for long-range Pedestrian and Cyclist detection under sparse observations.

\begin{figure*}[!t]
    \centering
    \includegraphics[width=\textwidth]{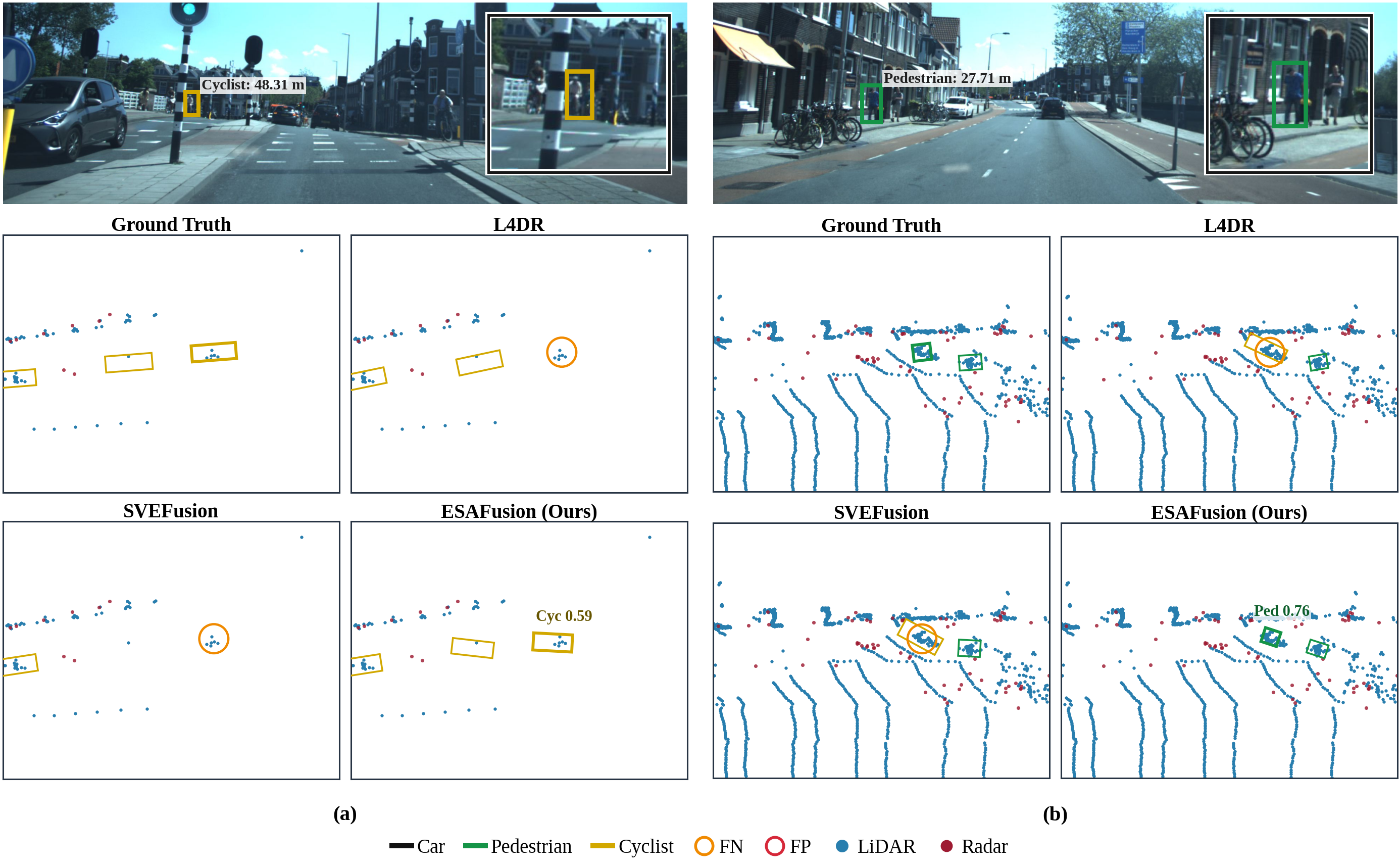}
    \caption{Distance-wise qualitative comparison on the VoD validation set.
    (a) ESAFusion detects a Cyclist at 48.31 m that is missed by L4DR and
    SVEFusion. (b) ESAFusion correctly classifies a Pedestrian at 27.71 m that
    is misclassified as a Cyclist by both baselines.}
    \label{fig:distance_qualitative}
\end{figure*}

Fig.~\ref{fig:distance_qualitative} illustrates representative challenging
cases in which ESAFusion recovers objects missed by the competing methods and
corrects category errors.

\begin{table*}[!t]
\centering
\caption{Quantitative results of different methods on the VoD-Fog dataset
using KITTI metrics under various fog levels. The best results are bold}
\label{tab:vod_fog_results}
\footnotesize
\renewcommand{\arraystretch}{1.02}
\begin{tabular*}{\textwidth}{@{\hspace{4pt}\extracolsep{\fill}} c l c *{9}{r} }
\toprule
\multirow{2}{*}{Fog Level} & \multirow{2}{*}{Methods}
& \multirow{2}{*}{Modality}
& \multicolumn{3}{c}{Car (IoU $=0.5$)}
& \multicolumn{3}{c}{Pedestrian (IoU $=0.25$)}
& \multicolumn{3}{c}{Cyclist (IoU $=0.25$)} \\
\cmidrule(lr){4-6}\cmidrule(lr){7-9}\cmidrule(l){10-12}
& & & Easy & Mod. & Hard & Easy & Mod. & Hard & Easy & Mod. & Hard \\
\midrule
\multirow{4}{*}{\shortstack{0\\(W/o Fog)}}
& PointPillars~\cite{lang2019pointpillars} & L   & 84.90 & 73.50 & 67.50 & 62.70 & 58.40 & 53.40 & 85.50 & 79.00 & 72.70 \\
& L4DR~\cite{huang2025l4dr}         & L+R & 85.00 & 76.60 & 69.40 & 74.40 & 72.30 & 65.70 & 93.40 & 90.40 & 83.00 \\
& SVEFusion~\cite{yang2026svefusion}    & L+R & \textbf{87.38} & 76.69 & 69.04 & \textbf{78.91} & 74.39 & 67.59 & 93.14 & 89.24 & 81.93 \\
& ESAFusion           & L+R & 86.82 & \textbf{76.87} & \textbf{69.84} & 78.69 & \textbf{74.43} & \textbf{68.27} & \textbf{95.26} & \textbf{90.45} & \textbf{83.35} \\
\midrule
\multirow{4}{*}{1}
& PointPillars~\cite{lang2019pointpillars} & L   & 79.90 & 72.70 & 67.00 & 59.90 & 55.60 & 50.50 & 85.50 & 78.20 & 72.00 \\
& L4DR~\cite{huang2025l4dr}         & L+R & 77.90 & 73.20 & 67.80 & 75.40 & 72.10 & 66.70 & \textbf{93.80} & \textbf{91.00} & 83.20 \\
& SVEFusion~\cite{yang2026svefusion}    & L+R & 83.41 & \textbf{75.32} & 66.97 & 75.82 & 73.35 & 67.31 & 92.70 & 89.02 & 82.60 \\
& ESAFusion           & L+R & \textbf{86.18} & 74.13 & \textbf{68.60} & \textbf{77.59} & \textbf{73.48} & \textbf{68.09} & 93.61 & 90.35 & \textbf{83.47} \\
\midrule
\multirow{4}{*}{2}
& PointPillars~\cite{lang2019pointpillars} & L   & 67.00 & 51.40 & 44.40 & 53.10 & 47.20 & 42.70 & 69.60 & 62.70 & 57.20 \\
& L4DR~\cite{huang2025l4dr}         & L+R & 68.50 & 56.40 & 49.30 & 63.10 & 59.90 & 55.10 & 82.70 & 70.80 & 70.70 \\
& SVEFusion~\cite{yang2026svefusion}    & L+R & 70.01 & 57.14 & 49.65 & 65.66 & 61.52 & 55.55 & 78.07 & 73.13 & 66.60 \\
& ESAFusion           & L+R & \textbf{73.37} & \textbf{59.40} & \textbf{52.15} & \textbf{66.59} & \textbf{63.09} & \textbf{56.56} & \textbf{84.62} & \textbf{79.53} & \textbf{72.32} \\
\midrule
\multirow{4}{*}{3}
& PointPillars~\cite{lang2019pointpillars} & L   & 44.50 & 31.90 & 27.00 & 40.20 & 37.70 & 34.00 & 53.20 & 46.70 & 41.80 \\
& L4DR~\cite{huang2025l4dr}         & L+R & 46.20 & 41.40 & 34.60 & 53.50 & 50.60 & 46.20 & 72.20 & 67.70 & 60.90 \\
& SVEFusion~\cite{yang2026svefusion}    & L+R & 50.19 & 39.38 & 34.09 & 54.11 & 51.36 & 45.47 & 64.66 & 60.26 & 55.05 \\
& ESAFusion           & L+R & \textbf{53.37} & \textbf{44.56} & \textbf{37.57} & \textbf{57.06} & \textbf{55.39} & \textbf{49.24} & \textbf{75.57} & \textbf{71.05} & \textbf{64.09} \\
\midrule
\multirow{4}{*}{4}
& PointPillars~\cite{lang2019pointpillars} & L   & 13.00 & 8.77 & 7.19 & 10.60 & 12.90 & 11.30 & 6.15 & 4.89 & 4.57 \\
& L4DR~\cite{huang2025l4dr}         & L+R & 26.90 & 26.20 & 21.60 & 33.10 & 30.70 & 27.90 & 30.30 & 29.70 & 26.30 \\
& SVEFusion~\cite{yang2026svefusion}    & L+R & 24.36 & 16.51 & 14.25 & 25.41 & 26.08 & 23.32 & 15.52 & 15.59 & 14.38 \\
& ESAFusion           & L+R & \textbf{33.17} & \textbf{31.08} & \textbf{26.32} & \textbf{35.13} & \textbf{34.56} & \textbf{30.79} & \textbf{38.62} & \textbf{36.90} & \textbf{33.34} \\
\bottomrule
\end{tabular*}
\end{table*}

\begin{figure*}[!t]
    \centering
    \includegraphics[
        width=\textwidth
    ]{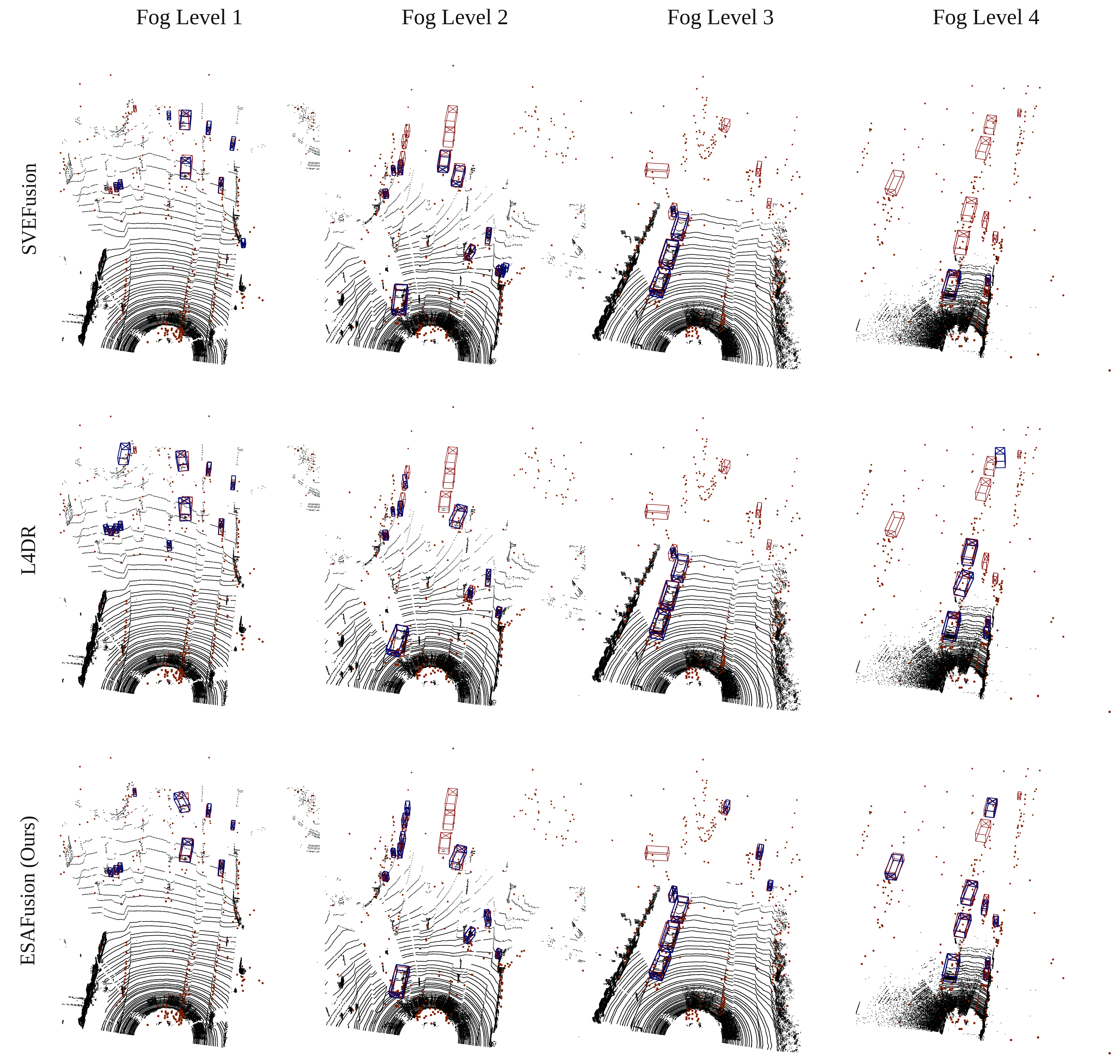}
    \caption{Qualitative 3-D object detection results on the VoD-Fog validation
    set under different simulated fog levels. Gray point clouds represent
    LiDAR data, while orange points represent 4-D radar data. Red boxes indicate
    ground-truth objects, and blue boxes indicate predicted bounding boxes.}
    \label{fig:vod_fog_qualitative}
\end{figure*}

\noindent\textbf{Robustness Evaluation on VoD-Fog}.
We evaluate our model in comparison with LiDAR-only methods and LiDAR--4-D
radar fusion methods on the VoD-Fog dataset using KITTI metrics across
varying fog levels.
As shown in Table~\ref{tab:vod_fog_results}, ESAFusion demonstrates greater robustness when
LiDAR observations are severely degraded.
Particularly under the most severe fog condition ($L=4$), our ESAFusion
method achieves performance improvements of 22.31\%, 21.66\%, and 32.01\%
AP over the LiDAR-only PointPillars~\cite{lang2019pointpillars} in moderate difficulty categories,
surpassing the gains obtained by L4DR~\cite{huang2025l4dr} and SVEFusion~\cite{yang2026svefusion}.
Compared with L4DR, ESAFusion further improves Cyclist AP by
8.32\%, 7.20\%, and 7.04\% under the Easy, Moderate, and Hard settings,
respectively.
These results highlight the effectiveness of radar evidence enhancement
and adaptive fusion under severe LiDAR degradation.
Fig.~\ref{fig:vod_fog_qualitative} further illustrates that ESAFusion maintains more reliable detections than L4DR and SVEFusion under progressively degraded LiDAR observations, with the advantage becoming particularly evident at Fog Levels 3 and 4.  

\subsection{Ablation Studies}

\noindent\textbf{Effect of Individual Modules}.
Table~\ref{tab:module_ablation} evaluates the individual contributions of
ERS, PCE, and ISAF. Each module independently improves detection performance.
Specifically, ERS first enhances point-level radar evidence and then
effectively filters out noisy 4-D radar points, while PCE facilitates the
fusion of LiDAR and 4-D radar pillar features by enhancing compact radar
evidence and introducing geometrically consistent local LiDAR support,
thereby improving the model's ability to exploit cross-modal
complementarity.
Building upon them, ISAF adaptively balances modality-specific
contributions and recalibrates the spatial responses of multiscale
features in the BEV backend, further improving mAP by 1.03\% and 2.49\%
in the Entire Area and RoI, respectively.
With all three modules enabled, the full model achieves the best overall
performance.

\begin{table}[!ht]
\centering
\caption{Ablation of different modules on the VoD validation set}
\label{tab:module_ablation}
\footnotesize
\setlength{\tabcolsep}{1.0pt}
\renewcommand{\arraystretch}{1.08}
\begin{tabular*}{\columnwidth}{@{\hspace{2pt}\extracolsep{\fill}} c c c *{8}{c} }
\toprule
\multicolumn{3}{c}{Module}
& \multicolumn{4}{c}{EAA}
& \multicolumn{4}{c}{RoI} \\
\cmidrule(lr){1-3}\cmidrule(lr){4-7}\cmidrule(l){8-11}
ERS & PCE & ISAF & Car & Ped. & Cyc. & mAP & Car & Ped. & Cyc. & mAP \\
\midrule
-- & -- & -- & 66.20 & 55.99 & 75.87 & 66.02 & 88.70 & 69.40 & 88.30 & 82.13 \\
\checkmark & -- & -- & 66.72 & 58.80 & 78.10 & 67.87 & 89.55 & 72.89 & 89.07 & 83.84 \\
-- & \checkmark & -- & 69.60 & 67.18 & 80.47 & 72.41 & 90.78 & 76.13 & 90.08 & 85.66 \\
\checkmark & \checkmark & -- & 70.25 & 68.54 & 81.92 & 73.57 & 90.90 & 77.83 & 90.48 & 86.40 \\
-- & -- & \checkmark & 69.40 & 67.55 & 81.80 & 72.92 & 90.75 & 76.95 & 93.75 & 87.15 \\
\checkmark & \checkmark & \checkmark
& \textbf{70.43} & \textbf{69.24} & \textbf{84.12} & \textbf{74.60}
& \textbf{90.92} & \textbf{78.59} & \textbf{97.16} & \textbf{88.89} \\
\bottomrule
\end{tabular*}
\end{table}

\begin{table}[!ht]
\centering
\caption{Ablation study of the PCE module on the VoD validation set}
\label{tab:pce_ablation}
\small
\renewcommand{\arraystretch}{1.08}
\begin{tabular*}{\columnwidth}{@{\hspace{8pt}\extracolsep{\fill}} c c c c c }
\toprule
CMC & AGI & CREE & EAA mAP & RoI mAP \\
\midrule
\checkmark & -- & -- & 73.02 & 86.71 \\
\checkmark & -- & \checkmark & 73.55 & 87.28 \\
\checkmark & \checkmark & -- & 74.01 & 88.35 \\
\checkmark & \checkmark & \checkmark & \textbf{74.60} & \textbf{88.89} \\
\bottomrule
\end{tabular*}
\end{table}

\begin{table}[!ht]
\centering
\caption{Ablation study of the neighborhood settings for AGI on the VoD validation set}
\label{tab:agi_neighborhood_ablation}
\small
\renewcommand{\arraystretch}{1.08}
\begin{tabular*}{\columnwidth}{@{\hspace{12pt}\extracolsep{\fill}} c c c }
\toprule
Range & EAA mAP & RoI mAP \\
\midrule
off         & 73.55 & 87.28 \\
4-connected & 74.12 & 88.52 \\
$3\times3$  & \textbf{74.60} & \textbf{88.89} \\
$5\times5$  & 73.90 & 87.84 \\
\bottomrule
\end{tabular*}
\end{table}

\noindent\textbf{Analysis of Pillar-Level Complementary Encoder.}
Table~\ref{tab:pce_ablation} evaluates the contributions of AGI and CREE while retaining CMC.
Both modules independently improve detection performance, with AGI
yielding larger gains than CREE, especially in the RoI.
When AGI and CREE are jointly introduced, mAP improves over the CMC
baseline by 1.58\% and 2.18\% in the EAA and RoI, respectively, while
CREE remains beneficial when AGI is enabled.
These results indicate that compact radar evidence and local LiDAR
support jointly strengthen radar pillar representations, enabling more
effective exploitation of cross-modal complementarity.

Table~\ref{tab:agi_neighborhood_ablation} compares different neighborhood settings for AGI.
Compared with disabling AGI, introducing neighboring LiDAR support
improves detection performance, indicating that radar pillars without
co-located LiDAR observations can benefit from local geometric
information.
The $3\times3$ neighborhood achieves the best performance in both
evaluation regions, whereas expanding the support range to $5\times5$
degrades performance.
This suggests that local complementation is not simply a matter of
incorporating more LiDAR features, as an excessively large neighborhood
may introduce irrelevant or weakly related geometric information from
the background or adjacent objects.

\begin{table}[!ht]
\centering
\caption{Ablation study of the ISAF module on the VoD validation set}
\label{tab:isaf_ablation}
\small
\renewcommand{\arraystretch}{1.08}
\begin{tabular*}{\columnwidth}{@{\hspace{12pt}\extracolsep{\fill}} l c c }
\toprule
Setting & EAA mAP & RoI mAP \\
\midrule
w/o ISAF                      & 73.57 & 86.40 \\
w/o Intra-Scale Gating        & 72.16 & 86.01 \\
w/o Inter-Scale Recalibration & 74.20 & 87.64 \\
Full ISAF                     & \textbf{74.60} & \textbf{88.89} \\
\bottomrule
\end{tabular*}
\end{table}

\noindent\textbf{Analysis of Intra- and Inter-Scale Adaptive Fusion.}
Table~\ref{tab:isaf_ablation} analyzes the roles of the two stages within ISAF.
Without intra-scale gating, retaining only basic multiscale
concatenation and alignment followed by inter-scale recalibration reduces
mAP by 1.41\% and 0.39\% in the Entire Area and RoI,
respectively, compared with the baseline without ISAF.
This suggests that applying region-adaptive scale recalibration directly
to concatenated multiscale features, without prior intra-scale gating,
may amplify unreliable responses.
In contrast, intra-scale gating alone improves mAP by 0.63\% and 1.24\%,
indicating that it effectively regulates LiDAR and radar feature
responses.
Combining intra-scale gating with inter-scale recalibration yields the best
performance. These results indicate that the two stages play complementary
roles: intra-scale gating regulates modality-specific feature responses at
each scale, whereas inter-scale recalibration adjusts the contributions of
different scales across BEV regions.
We further report computational complexity in giga floating-point
operations (GFLOPs), inference latency, and parameter count.

\begin{table}[!ht]
\centering
\caption{Ablation study of GFLOPs, latency, and parameter count for different
module configurations}
\label{tab:component_efficiency}
\footnotesize
\setlength{\tabcolsep}{2.2pt}
\renewcommand{\arraystretch}{1.08}
\begin{tabular}{@{}ccc ccc@{}}
\toprule
ERS & PCE & ISAF & GFLOPs & Latency (ms) & Params (M) \\
\midrule
-- & -- & -- & 88.19 & 13.55 & 18.05 \\
$\checkmark$ & -- & -- & 91.85 (+3.66) & 18.42 (+4.87) & 20.94 (+2.89) \\
$\checkmark$ & $\checkmark$ & -- & 91.90 (+0.05) & 43.34 (+24.92) & 21.16 (+0.22) \\
$\checkmark$ & $\checkmark$ & $\checkmark$ & 255.67 (+163.77) & 52.01 (+8.67) & 62.07 (+40.91) \\
\bottomrule
\end{tabular}
\end{table}

\noindent\textbf{Component-wise Efficiency Analysis.}
Table~\ref{tab:component_efficiency} summarizes the computational cost and
inference efficiency of the cumulative module configurations. ERS introduces
moderate additional overhead due to point-level radar evidence modeling and
foreground scoring and filtering with PointNet++. Although PCE adds little
arithmetic complexity, its sparse neighborhood operations incur a more
noticeable runtime overhead. ISAF produces the largest increases in
computation and parameter count because it processes multiscale BEV features.
Overall, the full ESAFusion maintains real-time inference, with an average
inference latency of 52.01~ms, corresponding to 19.23~FPS.

\section{Conclusion}
\label{sec:conclusion}

In this paper, we presented ESAFusion, an evidence-aware and scale-adaptive framework for LiDAR--4-D radar fusion that addresses challenges arising from sparse and unreliable radar observations, mismatched spatial sampling, and spatial variation in modality and scale importance.
ERS uses motion and observation-quality evidence to refine radar observations while preserving foreground confidence.
PCE performs co-located attribute complementation, extends LiDAR-to-radar geometric complementation beyond strict co-location using neighboring LiDAR support based on geometric consistency, and enhances radar pillars with compact radar evidence.
ISAF realizes multiscale adaptive interaction through evidence-refined intra-scale modality gating and support-conditioned inter-scale recalibration.
Extensive experiments demonstrate that ESAFusion achieves strong detection performance on VoD and Astyx datasets, while supporting real-time inference and maintaining robustness under progressively degraded LiDAR observations on VoD-Fog.
Future work will explore efficient temporal modeling for LiDAR--4-D radar fusion in complex driving scenarios.

\bibliographystyle{IEEEtran}
\bibliography{references}

% \newpage

\vspace{-1.cm}
\begin{IEEEbiography}[{\includegraphics[width=1in,height=1.25in,clip,keepaspectratio]{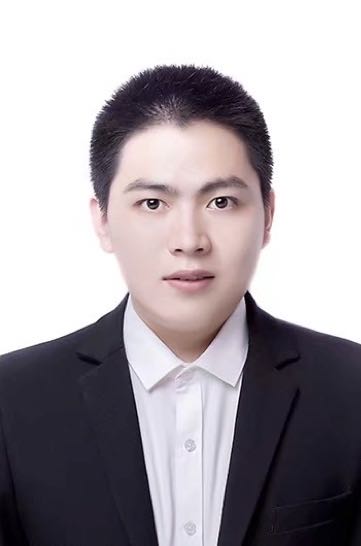}}]
{Gang Ma} received his undergraduate Bachelor's degree in computer science and technology from Soochow University (Suzhou, China) in 2018. He then completed his Ph.D in the School of Computer Science, Fudan University (Shanghai, China) in 2022. Currently, he is an assistant professor at the School of Future Technology, Shanghai University (Shanghai, China).

His current research interests include computer vision, robotics, artificial intelligence and intelligent unmanned systems.
\end{IEEEbiography}

% \vspace{-1.cm}
\begin{IEEEbiography}[{\includegraphics[width=1in,height=1.25in,clip,keepaspectratio]{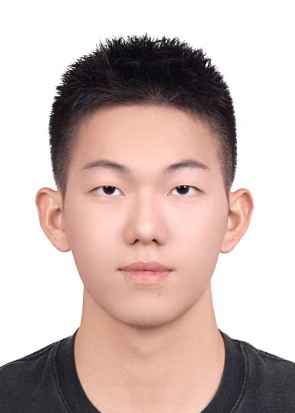}}]
{Senjie Hu} received the B.S. degree in Computer Science and Technology from Zhejiang Agriculture and Forestry University, Hangzhou, China. He is currently pursuing the Master's degree in Artificial Intelligence with the School of Future Technology, Shanghai University, Shanghai, China. 

His research interests include 3D object detection, multi-modal sensor fusion (such as LiDAR, camera, and 4D imaging radar), and autonomous perception for Unmanned Surface Vehicles (USVs).
\end{IEEEbiography}

\vspace{-16.cm}
\begin{IEEEbiography}[{\includegraphics[width=1in,height=1.25in,clip,keepaspectratio]{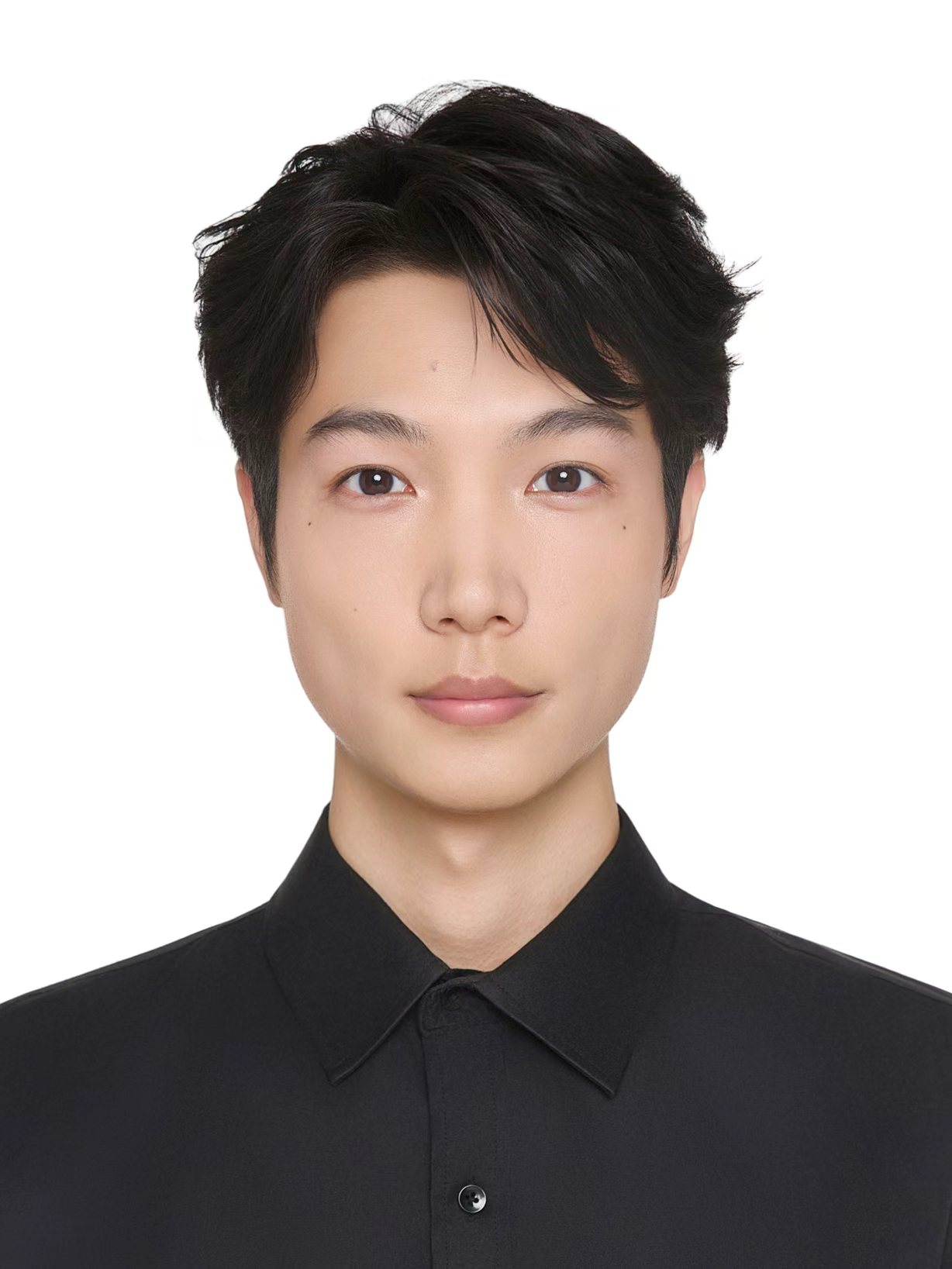}}]
{Junjie Liu} received the B.S. degree in Network Engineering from Ningbo University of Technology, Ningbo, China. He is currently pursuing the M.S. degree in Artificial Intelligence with the School of Future Technology, Shanghai University, Shanghai, China.

His research interests include multi-object tracking, intent recognition, and autonomous perception for Unmanned Surface Vehicles (USVs).
\end{IEEEbiography}

\newpage
% \vspace{-2.cm}
\begin{IEEEbiography}[{\includegraphics[width=1in,height=1.25in,clip,keepaspectratio]{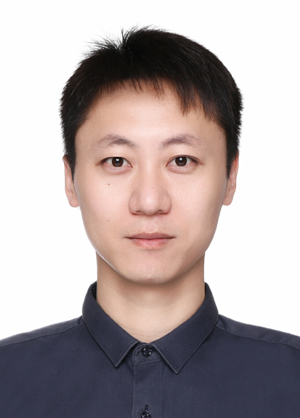}}]
{Chao Wang} received his Ph.D. degree from the School of Computer Science, Fudan University in 2022. He is an assistant professor at the School of Future Technology, Shanghai University. 

His research interests include natural language processing, knowledge bases, and causal inference.
\end{IEEEbiography}

\vspace{-16.0cm}
\begin{IEEEbiography}[{\includegraphics[width=1in,height=1.25in,clip,keepaspectratio]{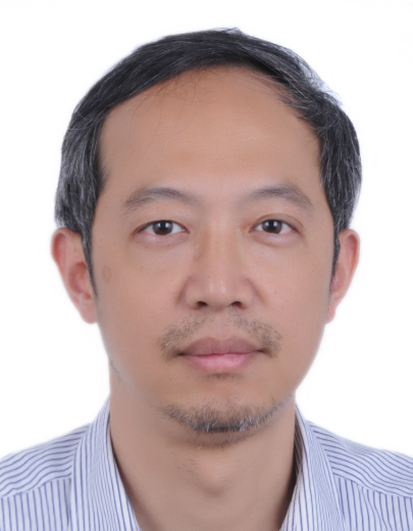}}]
{Hui Wei} received the Ph.D. degree from the Department of Computer Science, Beijing University of Aeronautics and Astronautics, in 1998. From 1998 to 2000, he was a Post-Doctoral Fellow with the Department of Computer Science and the Institute of Artificial Intelligence, Zhejiang University. Since November 2000, he has been with the Department of Computer Science and Engineering, Fudan University. 

His research interests include artificial intelligence and cognitive science.
\end{IEEEbiography}

% \section{Biography Section}
% If you have an EPS/PDF photo (graphicx package needed), extra braces are
%  needed around the contents of the optional argument to biography to prevent
%  the LaTeX parser from getting confused when it sees the complicated
%  $\backslash${\tt{includegraphics}} command within an optional argument. (You can create
%  your own custom macro containing the $\backslash${\tt{includegraphics}} command to make things
%  simpler here.)
 
% \vspace{11pt}

% \bf{If you include a photo:}\vspace{-33pt}
% \begin{IEEEbiography}[{\includegraphics[width=1in,height=1.25in,clip,keepaspectratio]{fig1}}]{Michael Shell}
% Use $\backslash${\tt{begin\{IEEEbiography\}}} and then for the 1st argument use $\backslash${\tt{includegraphics}} to declare and link the author photo.
% Use the author name as the 3rd argument followed by the biography text.
% \end{IEEEbiography}

% \vspace{11pt}

% \bf{If you will not include a photo:}\vspace{-33pt}
% \begin{IEEEbiographynophoto}{John Doe}
% Use $\backslash${\tt{begin\{IEEEbiographynophoto\}}} and the author name as the argument followed by the biography text.
% \end{IEEEbiographynophoto}

% \vfill

\end{document}